\documentclass{article}

\usepackage[final]{corl_2026}
\usepackage{amsmath}
\usepackage{amssymb}
\usepackage{booktabs}
\usepackage{placeins}
\usepackage{caption}
\usepackage{threeparttable}
\usepackage{xcolor}
\usepackage{enumitem}
\usepackage{fontawesome5}
\usepackage{xspace}
\usepackage{graphicx}
\usepackage{tabularx}
\usepackage{fancyvrb}
\usepackage{pifont}
\usepackage{float}
\usepackage{algorithm}
\usepackage{algorithmic}
\newcommand{\cmark}{\checkmark}
\newcommand{\xmark}{\ding{55}}

\usepackage[most]{tcolorbox}
\usepackage{ragged2e}

\definecolor{promptbg}{HTML}{F8FAF7}
\definecolor{promptframe}{HTML}{CBD5C0}
\definecolor{prompttitle}{HTML}{EEF4EA}

\newtcolorbox{promptbox}{
  enhanced,
  breakable,
  colback=promptbg,
  colframe=promptframe,
  coltitle=black,
  colbacktitle=prompttitle,
  title={Background inpainting prompt},
  fonttitle=\bfseries\small,
  boxrule=0.45pt,
  arc=1.5mm,
  left=2.5mm,
  right=2.5mm,
  top=1.5mm,
  bottom=1.5mm,
  before upper={
    \RaggedRight
    \small
    \setlength{\parindent}{0pt}
    \setlength{\parskip}{0.55em}
    \hyphenpenalty=10000
    \exhyphenpenalty=10000
    \sloppy
  },
}

\newtcblisting{yamlbox}{
  enhanced,
  breakable,
  listing only,
  colback=promptbg,
  colframe=promptframe,
  coltitle=black,
  colbacktitle=prompttitle,
  title={Example task-generation YAML},
  fonttitle=\bfseries\small,
  boxrule=0.45pt,
  arc=1.5mm,
  left=2.5mm,
  right=2.5mm,
  top=1.5mm,
  bottom=1.5mm,
  listing options={
    basicstyle=\ttfamily\scriptsize,
    columns=fullflexible,
    breaklines=true,
    keepspaces=true,
    showstringspaces=false
  }
}

\definecolor{robosnapshadow}{HTML}{6F8F4E}
\definecolor{robosnapyellow}{HTML}{FFDA67}

\newcommand{\method}{\textsc{RoboSnap}\xspace}

\newcommand{\methodiconshadow}{%
  \begingroup
  \setlength{\fboxsep}{0pt}%
  \ooalign{%
  \hskip-1.4pt\raisebox{0pt}{%
      \textcolor{robosnapshadow}{%
        \scalebox{0.6}{\faCameraRetro}\,\textsc{RoboSnap}%
      }%
    }\cr
    \hskip-1.2pt\raisebox{0pt}{%
      \textcolor{robosnapshadow}{%
        \scalebox{0.6}{\faCameraRetro}\,\textsc{RoboSnap}%
      }%
    }\cr
    \hskip-1.0pt\raisebox{0pt}{%
      \textcolor{robosnapshadow}{%
        \scalebox{0.6}{\faCameraRetro}\,\textsc{RoboSnap}%
      }%
    }\cr
    \hskip-0.7pt\raisebox{0pt}{%
      \textcolor{robosnapshadow}{%
        \scalebox{0.6}{\faCameraRetro}\,\textsc{RoboSnap}%
      }%
    }\cr
    \textcolor{robosnapyellow}{%
      \scalebox{0.6}{\faCameraRetro}\,\textsc{RoboSnap}%
    }\cr
  }%
  \endgroup\xspace%
}

\newcommand{\robosnapshadowtext}[1]{%
  \begingroup
  \setlength{\fboxsep}{0pt}%
  \ooalign{%
  \hskip-0.7pt\raisebox{0pt}{%
      \textcolor{robosnapshadow}{#1}%
    }\cr
    \hskip-0.6pt\raisebox{0pt}{%
      \textcolor{robosnapshadow}{#1}%
    }\cr
    \hskip-0.5pt\raisebox{0pt}{%
      \textcolor{robosnapshadow}{#1}%
    }\cr
    \hskip-0.4pt\raisebox{0pt}{%
      \textcolor{robosnapshadow}{#1}%
    }\cr
    \textcolor{robosnapyellow}{#1}\cr
  }%
  \endgroup%
}

\usepackage{multirow}
\usepackage{hyperref}
\usepackage{subcaption}
\usepackage{wrapfig}
\usepackage{pgfplots}

\newcommand{\dataset}{DROID-Sim\xspace}

\title{
{\fontsize{17.5}{21}\selectfont
\methodiconshadow: One-Shot Real-to-Sim Scene Generation for Generalizable Robot Learning and Evaluation
\par\vspace{0.6em}
}
}

\author{
\makebox[\textwidth][c]{%
\begin{minipage}{0.98\textwidth}
\centering
\textbf{
Shujie Zhang\textsuperscript{\normalfont 1\,4\,*} \quad
Jingkun Yi\textsuperscript{\normalfont 1\,3\,*} \quad
Weipeng Zhong\textsuperscript{\normalfont 1\,2} \quad
Zirui Zhou\textsuperscript{\normalfont 4} \quad
Yangkun Zhu\textsuperscript{\normalfont 1}
}\\[0.25em]
\textbf{
Hanqing Wang\textsuperscript{\normalfont 1} \quad
Xudong Xu\textsuperscript{\normalfont 1\,\dag} \quad
Weinan Zhang\textsuperscript{\normalfont 1\,2} \quad
Chunhua Shen\textsuperscript{\normalfont 1\,3}
}\\[0.55em]
{\normalfont\footnotesize
\textsuperscript{1}Shanghai AI Laboratory \quad
\textsuperscript{2}Shanghai Jiao Tong University \quad
\textsuperscript{3}Zhejiang University
}\\[0.05em]
{\normalfont\footnotesize
\textsuperscript{4}Tsinghua University
}\\[0.35em]
{\normalfont\footnotesize
\robosnapshadowtext{Homepage:}~
\href{https://robosnap.github.io/}{%
  \textcolor{robosnapshadow}{\texttt{https://robosnap.github.io}}%
}
}
\end{minipage}%
}
}

\begin{document}
\maketitle
\begingroup
\makeatletter
\renewcommand{\@makefntext}[1]{\noindent #1}
\makeatother
\renewcommand{\thefootnote}{}
\footnotetext{%
\begin{tabular}{@{}l@{}}
\textsuperscript{*}Equal contribution. \quad
\textsuperscript{\dag}Corresponding author.\\
\end{tabular}%
}
\endgroup

\vspace{-2.2em}
\begin{figure}[H]
  \centering
  \includegraphics[width=0.9\linewidth]{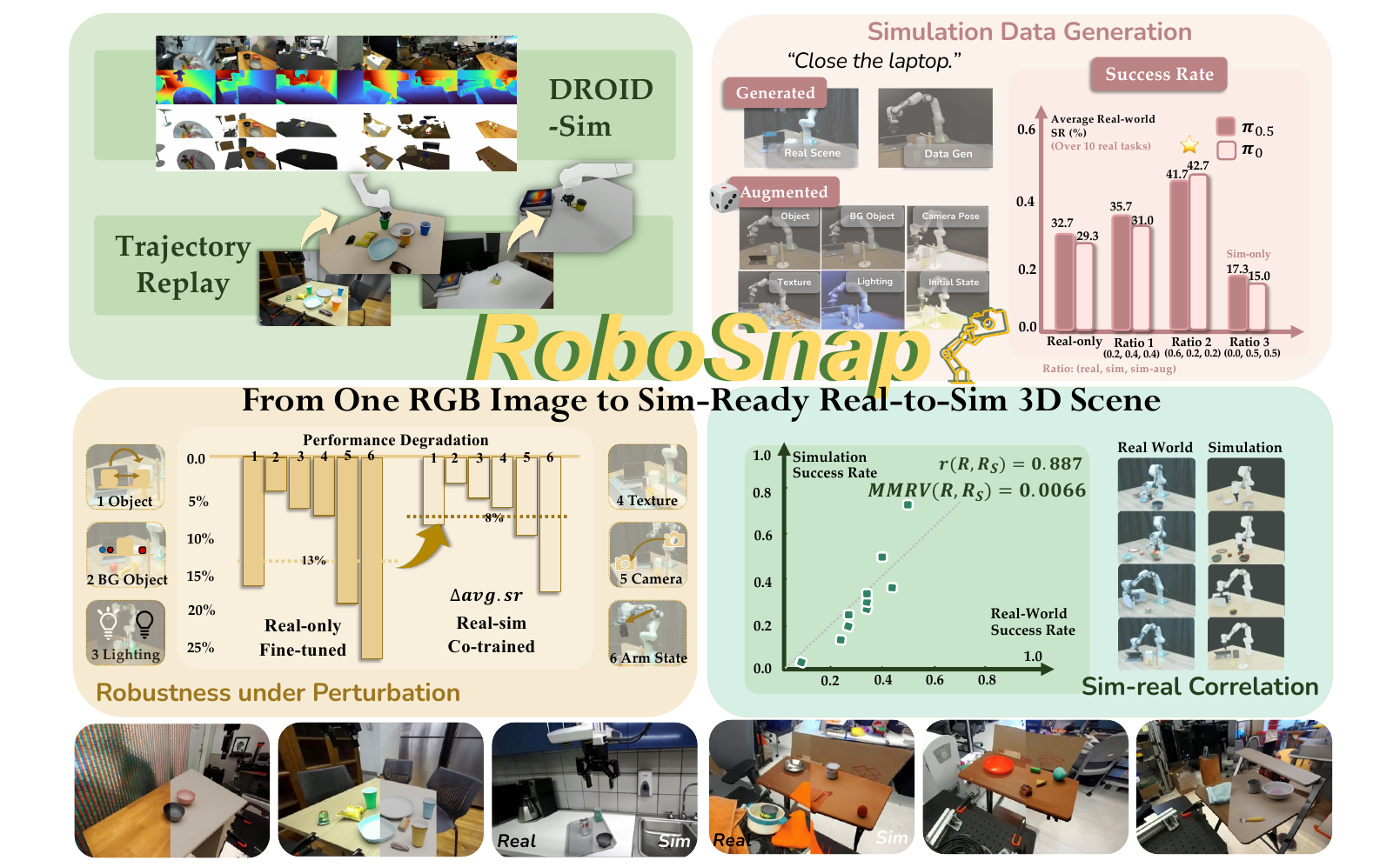}
  \caption{From a single RGB image, \method reconstructs a reusable simulation-ready scene with interactive physical assets and visual context. The recovered scenes support trajectory replay (\textbf{top-left}), task-specific data generation and augmentation (\textbf{top-right}, \textbf{bottom-left}), and policy evaluation with meaningful sim-real correlation (\textbf{bottom-right}).}
  \label{fig:teaser}
\vspace{-0.5em}
\end{figure}

\begin{abstract}
Recovering real-world scenes as interactive simulation environments can enable generalizable robot learning and reproducible policy evaluation. However, constructing scenes that are both physically stable and visually faithful remains slow and expensive. In this work, we present \method, a real-to-sim framework that turns a single RGB image into a simulation-ready scene. The key idea is a layered design that separates the physics-critical interaction area from the surrounding visual context: collision-aware foreground assets are refined for stable robot interaction, while a 3D Gaussian splatting visual layer preserves faithful background appearance under novel views. Experiments on DROID scenes and real-robot tasks show that \method achieves reliable trajectory replay in the recovered scenes, supports task-specific synthetic data generation for policy training, and yields meaningful sim-real correlation for policy evaluation. To further support real-to-sim research, we introduce \dataset, a real-to-sim companion dataset constructed from 564 real-world scenes in DROID. Extensive experiments suggest that the value of real-to-sim methods lies not only in high-fidelity visual reconstruction, but in turning real environments into reusable infrastructure for robot learning and evaluation.

\end{abstract}

\keywords{Real-to-Sim-to-Real, Robot Data Generation, Vision-Language-Action Models, Robot Manipulation}
\section{Introduction}
\label{sec:introduction}

Recent robot foundation models formulate manipulation as a conditional action generation task from visual, linguistic, and proprioceptive inputs~\citep{rt1,octo,openvla2024,pi02024,pi05}. As these models scale, large-scale training data and reproducible evaluation have become critical bottlenecks. Although real-world datasets and benchmarks provide substantial physical demonstrations and standardized protocols~\citep{droid2024,bridgev2,oxe2023,maniskill2}, scalable data acquisition and flexible policy evaluation remain costly, labor-intensive, and hardware-bound. Simulation offers a complementary path for scalable data synthesis, scene augmentation, and repeatable policy assessment, thereby motivating the construction of interactive simulation scenes that are both physically plausible and visually faithful to real-world deployment environments.

Existing approaches only partially satisfy these requirements. Procedural and generative scene synthesis methods have scaled simulatable environments for robot learning~\citep{infinigen2023infinite,mesatask2025,yang2024physcene,wang2025tabletopgen, choi2026scalingsimtorealreinforcementlearning}, but they primarily focus on creating diverse simulation scenes rather than recovering reusable interactive replicas of specific in-the-wild real-world scenes. Reconstruction-based real-to-sim methods improve scene alignment but often require multi-view capture or manual refinement~\citep{rialto2024,re3sim2025,splatsim2024,yu2025metascenes,polaris2025}. Recent single-image systems reduce the capture burden~\citep{acdc2024, yao2025cast, rola2025}, but their outputs typically target narrower endpoints: retrieval-based digital cousins, task or demonstration synthesis, or partially recovered scenes with static background. As a result, they do not generally recover persistent simulation worlds that can be re-rendered, edited, and reused from new viewpoints, and their effectiveness in downstream robot learning and evaluation workflows remains underexplored. Motivated by recent advances in monocular 3D geometry and image-conditioned 3D asset generation~\citep{vggt2025, wang2026vggtomega, sam3d2025, trellis2025}, we ask: \emph{Can we reconstruct a physically plausible, visually faithful, and simulation-ready environment from a single RGB image?}

To this end, we present \method, a single-image real-to-sim method for manipulation scene generation. \method reconstructs the interaction area as collision-aware objects and support surfaces, aligns the area to a gravity-consistent frame, and refines object poses to resolve floating artifacts, interpenetrations, and unstable contacts. The surrounding context is modeled as a separate visual layer using background completion, Gaussian splatting, and scene lighting. The resulting scene is editable, reusable, and simulation-ready for robot data generation and evaluation.

We make the following contributions. (i)~We propose \method, a layered real-to-sim method that converts a single RGB image into a simulation-ready scene. (ii)~We systematically validate that the resulting scenes support downstream robot-learning workflows, including trajectory replay, data generation, and policy evaluation with meaningful correlation to real-world performance. (iii)~We introduce \dataset, a real-to-sim companion dataset of 564 DROID scenes~\citep{droid2024}, extending an existing robot dataset from recorded trajectories and images to reusable simulation environments.
\section{Related Work}
\label{sec:related}

\paragraph{3D Generation and Scene Synthesis.}
Procedural and generative scene-construction systems have scaled simulatable environments for robot learning, from everyday manipulation scenes to physically interactable tabletop and indoor layouts~\citep{robocasa2024,robocasa365,mesatask2025,yang2024physcene,yang2025sceneweaver,internscenes2025}.
However, they typically synthesize new environments rather than reconstructing site-specific real-world robot scenes.
Meanwhile, advances in monocular geometry and image-conditioned 3D asset generation enable 3D structure inference from limited visual inputs~\citep{depth_pro2024,vggt2025,sam3d2025,trellis2025,hunyuan3d2025}; \method leverages these advances for real-to-sim, converting a single image into a physically plausible and visually faithful simulation environment.

\paragraph{Real-to-Sim-to-Real.}
Real-to-sim methods reconstruct digital twins or task-specific simulation assets from real observations for policy learning and evaluation~\citep{rialto2024,re3sim2025,splatsim2024,robogs2024,twinaligner2025,polaris2025,yu2025metascenes,scalable_real2sim2025}. Their alignment gains, however, typically come from additional information or intervention beyond a single RGB image, making such methods less lightweight and harder to reuse across different scenes. Recent single-observation methods reduce the capture burden~\citep{acdc2024,grs2025,yao2025cast,rola2025}, but they often target narrower endpoints such as asset retrieval, static reconstruction, or demonstration synthesis, rather than producing reusable simulation worlds and validating the recovered scenes in downstream robot-learning workflows. \method instead reconstructs a single RGB image as a layered, physically refined manipulation scene that can be re-rendered, edited, and reused from new viewpoints.

\paragraph{Robot Data Generation and Policy Evaluation.}
Large robot datasets provide foundational data for scalable training and reproducible evaluation~\citep{droid2024,bridgev2,oxe2023,maniskill2}, 
while generalist policies further motivate standardized benchmarks across diverse tasks~\citep{rt1,octo,openvla2024,pi02024,pi05}.
Recent work has expanded robot datasets using generative visual methods~\citep{fang2025rebot,roboengine2025,real2render2real2025,robosplat,wang2026robovip,real2edit2real2025} 
or simulation- and task-construction approaches~\citep{re3sim2025,mimicgen2023,dexmimicgen2024,robogen2024,tian2025interndata}.
Several simulation benchmarks provide standardized tasks and environments for robot evaluation~\citep{maniskill2,liu2023libero,behavior1k,chen2025robotwin,robocasa2024,robocasa365,simplerenv2024,polaris2025}, and well-constructed simulated scenes can yield meaningful correlations with real-world policy performance. \method complements these efforts by converting real in-the-wild scene images, whether from robot datasets or casual captures, into reusable, simulation-ready environments for trajectory-based data generation and flexible policy evaluation.

\section{Method}
\label{sec:method}

\begin{figure}[htbp]
    \centering
    \includegraphics[width=\linewidth]{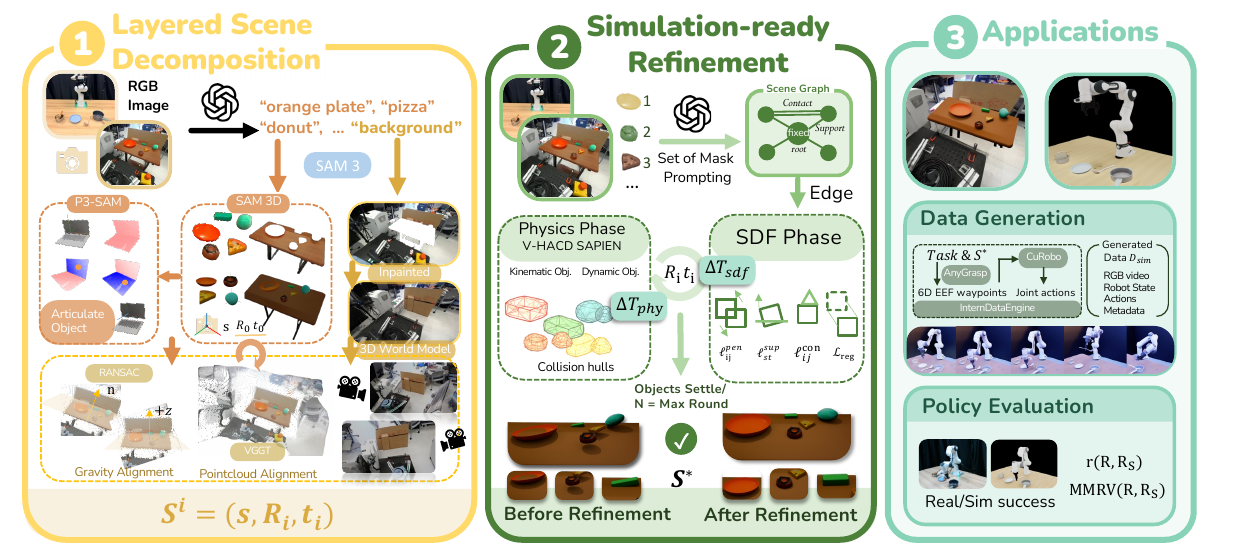}
    \caption{\textbf{Overview of \method.}
(1)~From a single RGB image, \method decomposes the scene into an interactive physical layer and a re-renderable visual context layer. (2)~The resulting layered scene is refined to resolve severe physical instabilities. (3)~The simulation-ready scene supports task-specific synthetic data generation and closed-loop policy evaluation.
}
    \label{fig:pipeline}
    \vspace{-1.0em}
\end{figure}

\subsection{Layered Scene Reconstruction from a Single Image}
\label{sec:method_1}

\paragraph{Problem formulation.}
Given a single RGB image $I\in\mathbb{R}^{H_0\times W_0\times 3}$, \method first reconstructs an initial layered scene $\mathcal{S}^{(0)}$ and then refines it into a simulation-ready scene $\mathcal{S}^{\star}$. The scene contains a \emph{physical layer} of interactable objects and support surfaces, and a \emph{visual layer} for the surrounding context. Both layers are registered to a canonical world frame $W$, whose origin is the support-platform centroid with $-\hat{\mathbf{e}}_z$ as the gravity direction.

\paragraph{Interactive Physical Layer.}
We use a VLM~\citep{achiam2023gpt} to parse the interaction area and identify object names $\{\ell_i\}_{i=1}^{N}$, including the support platform.
SAM~3~\citep{sam3_2025} extracts instance masks $\{M_i\}_{i=1}^{N}$, and SAM~3D~\citep{sam3d2025} reconstructs each object as a textured mesh $\mathcal{M}_i$ with an initial pose and scale. 
To further register these object assets to the scene geometry, we use VGGT~\citep{vggt2025} to predict camera geometry, confidence, and a dense point map $\mathcal{X}_V$ in the camera frame $V$. 
For each object mask, we extract high-confidence foreground points from the VGGT point map and refine the initial SAM~3D pose using mask-guided registration. 
In practice, we perform coarse-to-fine fixed-scale ICP between sampled mesh surface points and the corresponding foreground point cloud, and reject updates with excessive rotation or translation. 
The resulting aligned pose is denoted as $T_{i\to V}^{\mathrm{init}}$.

\paragraph{Canonical Alignment and Robot Base.}
We estimate $W$ from the dominant support platform by selecting support points from $\mathcal{X}_V$ and fitting a plane with RANSAC followed by least-squares refinement.
$T_{V\to W}$ aligns the plane normal to $\hat{\mathbf{e}}_z$ and places the support-platform centroid at the origin.
Object poses are lifted to $W$ by
\begin{equation}
T_{i\to W}^{\mathrm{init}} = T_{V\to W}T_{i\to V}^{\mathrm{init}} .
\label{eq:obj-to-W}
\end{equation}
For calibrated datasets such as DROID~\citep{droid2024}, the robot base is placed using the camera-frame base pose: $T_{B\to W}=T_{V\to W}T_{B\to V}.$
For uncalibrated captures, the robot base can be initialized from the support-platform geometry and a specified robot-facing direction.

\paragraph{Visual Context Layer.}
We reconstruct the non-interactable context separately.
After foreground masking, we inpaint missing regions using a VLM-guided prompt (Appendix~\ref{app:background}) and pass the completed image to a generative world model~\citep{marble2025}, producing a Gaussian-splat scene $\mathcal{G}_M$ in frame $F_M$.
We align $\mathcal{G}_M$ to $V$ using sparse correspondences and ICP, then lift it to $W$ with $T_{F_M\to W}=T_{V\to W}T_{F_M\to V}.$

\paragraph{Articulated Objects.}
For articulated objects, we decompose the reconstructed mesh into semantic parts using a point-based part segmentation model~\citep{p3sam2025}, split the mesh by estimated part boxes, and attach the recovered part meshes to category-level kinematic parameters retrieved from an articulated-object dataset~\citep{Xiang_2020_SAPIEN}.

Collecting these components, we denote the initial layered scene as
$\mathcal{S}^{(0)}=(\{(M_i,\ell_i,\mathcal{M}_i,T_{i\to W}^{\mathrm{init}})\}_{i=1}^{N},(\mathcal{G}_M,T_{F_M\to W}),W,T_{B\to W})$.

\subsection{Simulation-ready Refinement}
\label{sec:method-refine}

Independent per-object pose estimates often produce floating objects, interpenetrations, and unstable contacts. We therefore extract a physical scene graph and refine object poses with an alternating SDF--physics procedure: the SDF phase enforces support/contact constraints and resolves geometric conflicts, while the physics phase settles objects under gravity.

\paragraph{Scene Graph Extraction.}
Inspired by CAST~\citep{yao2025cast}, we infer pairwise physical relations with Set-of-Mark prompting~\citep{yang2023set} and a VLM~\citep{achiam2023gpt}. From the instance masks and captions in \S\ref{sec:method_1}, GPT-4V predicts relations over $K=5$ randomized SoM overlays. Majority-voted predictions define directed \emph{Support} edges and bidirectional \emph{Contact} edges, yielding $\mathcal{G}_{\mathrm{phys}}=(\mathcal{V},\mathcal{E}_{\mathrm{sup}}\cup\mathcal{E}_{\mathrm{con}})$. Objects that only support others are fixed as roots $\mathcal{R}$.

\paragraph{Alternating SDF--physics Optimization.}
Given $\mathcal{G}_{\mathrm{phys}}$ and initial poses $\{T_{i\to W}^{\mathrm{init}}\}$ from Eq.~\eqref{eq:obj-to-W}, we refine non-root object poses by optimizing residual SE(3) updates:
\begin{equation}
T_{i \to W}
=
\Delta T_i T_{i\to W}^{\mathrm{init}},
\qquad
\Delta T_i
=
\begin{bmatrix}
\exp([\Delta\mathbf{r}_i]_\times) & \Delta\mathbf{t}_i\\
\mathbf{0}^{\top} & 1
\end{bmatrix},
\quad
i\notin\mathcal{R}.
\label{eq:delta-pose}
\end{equation}

The refinement alternates between an SDF optimization phase and a physics settling phase. The SDF phase uses precomputed SDF grids and surface samples to minimize penetration, support, contact, and regularization losses, with full formulae in Appendix~\ref{app:loss}. We precompute V-HACD collision hulls~\citep{vhacd_collision} for the physical assets. The physics phase simulates these assets in SAPIEN~\citep{Xiang_2020_SAPIEN}, keeping root objects kinematic and other objects dynamic. The Gaussian background serves as a non-colliding visual layer and does not participate in physics refinement. The settled poses initialize the next SDF phase, and the final poses define $\mathcal{S}^{\star}$. The refined scenes are then integrated into Isaac Sim-based workflows for trajectory replay, data generation, and policy evaluation.

\subsection{Robot Data Generation and Evaluation}
\label{sec:method-render}

The refined scene $\mathcal{S}^{\star}$ supports trajectory replay, task-specific data generation, and closed-loop policy evaluation. For downstream robot applications, we use the camera--robot extrinsics provided by DROID dataset for trajectory replay and camera--robot calibration for real-robot deployment. 

\paragraph{Layered Rendering.}
For a query camera $Q$, Isaac Sim~\citep{isaacsim} renders the physical layer as $(I_{\mathrm{fg}},D_{\mathrm{fg}},\alpha_{\mathrm{fg}})$, while the visual layer is rendered from the same camera transformed to the Gaussian splatting frame, $T_{Q\to F_M}=T_{F_M\to W}^{-1}T_{Q\to W}$, yielding $(I_{\mathrm{bg}},D_{\mathrm{bg}})$. We depth-composite the two layers as
\begin{equation}
I_{\mathrm{out}}(\mathbf{u})
=
m(\mathbf{u})I_{\mathrm{fg}}(\mathbf{u})
+
(1-m(\mathbf{u}))I_{\mathrm{bg}}(\mathbf{u}),
\quad
m(\mathbf{u})=
\mathbf{1}\!\left[
\alpha_{\mathrm{fg}}(\mathbf{u})>0
\wedge
D_{\mathrm{fg}}(\mathbf{u})\le D_{\mathrm{bg}}(\mathbf{u})
\right].
\label{eq:compositing}
\end{equation}

\paragraph{Trajectory-based Data Generation.}
We instantiate $\mathcal{S}^{\star}$ in an Isaac Sim-based data engine~\citep{tian2025interndata} with a task specification, robot embodiment, query cameras, and the layered renderer. Grasp-centric skills use AnyGrasp-initialized candidates~\citep{anygrasp2023}; skill modules output target end-effector 6D waypoints, which cuRobo converts into collision-aware dense joint-space actions~\citep{curobo2026}. 

\paragraph{Policy Evaluation.}
For closed-loop evaluation, actions are executed in $\mathcal{S}^{\star}$. For sim-real evaluation, let $\mathbf{R}$ and $\mathbf{R}_{S}$ denote real and simulated success-rate vectors over $N$ tasks or checkpoints. We report Pearson correlation:

$$ r(\mathbf{R},\mathbf{R}_{S})=\frac{\sum_i (R_i-\bar{R})(R_{S,i}-\bar{R}_S)}{\sqrt{\sum_i (R_i-\bar{R})^2}\sqrt{\sum_i (R_{S,i}-\bar{R}_S)^2}}$$
to measure success-rate agreement, and mean maximum rank violation $$\mathrm{MMRV}(\mathbf{R},\mathbf{R}_S)=\frac{1}{N}\sum_i\max_j |R_i-R_j|\mathbf{1}[(R_{S,i}<R_{S,j})\neq(R_i<R_j)]$$
to measure rank-order inconsistency. Higher $r$ and lower MMRV~\citep{simplerenv2024} indicate better sim-real alignment.

\section{Experiments}
\label{sec:experiment}

We organize our experiments around five questions that probe what single-image real-to-sim scene recovery can support in robot learning and evaluation:

\begin{description}[leftmargin=0pt,labelsep=0.4em,itemsep=0pt,topsep=1pt,parsep=0pt,partopsep=0pt]
\item[\textbf{Q1: Visual realism and simulation readiness.}] Are the reconstructed scenes visually faithful to the input image and physically stable in simulation?
\item[\textbf{Q2: Trajectory replay.}] Do the recovered scenes preserve the geometry and contact structure needed to replay real robot trajectories?
\item[\textbf{Q3: Data generation and policy fine-tuning.}] Can \method-generated scenes improve real-world policy performance through task-specific data generation?
\item[\textbf{Q4: Robustness under perturbation.}] Are policies trained with \method-generated data robust to real-world perturbations?
\item[\textbf{Q5: Generative evaluation harness.}] Can \method scenes serve as a generative evaluation harness whose simulated rollouts correlate with real-world policy performance?
\end{description}

\subsection{Visual Realism and Simulation Stability}
\label{sec:exp-realism}

We construct \dataset{} by running \method{} on 564 DROID scenes. For detailed quantitative evaluation and baseline comparison, we select 10 scenes covering diverse room layouts, object counts ranging from 2 to 15 with varying task difficulty. Dataset and scene details are provided in Appendix~\ref{app:droid-sim}.

\begin{figure}[htbp]
\vspace{-0.5em}
\centering
\begin{minipage}{\linewidth}
\centering
\small
\setlength{\tabcolsep}{3.2pt}
\resizebox{0.95\linewidth}{!}{%
\begin{tabular}{lcccccc}
\toprule
Method & PSNR $\uparrow$ & SSIM $\uparrow$ & LPIPS $\downarrow$ & SIFT-MR $\uparrow$ & RGB-EMD $\downarrow$ & Gabor-L1 $\downarrow$ \\
\midrule
RoLA~\citep{rola2025} & 13.40 & 0.4521 & 0.4996 & 0.0664 & 28.5806 & 0.001817 \\
\textbf{\method (ours)} & 13.25 & \textbf{0.4907} & \textbf{0.4958} & \textbf{0.1226} & \textbf{11.4795} & \textbf{0.000741} \\
\bottomrule
\end{tabular}%
}
\includegraphics[width=\linewidth]{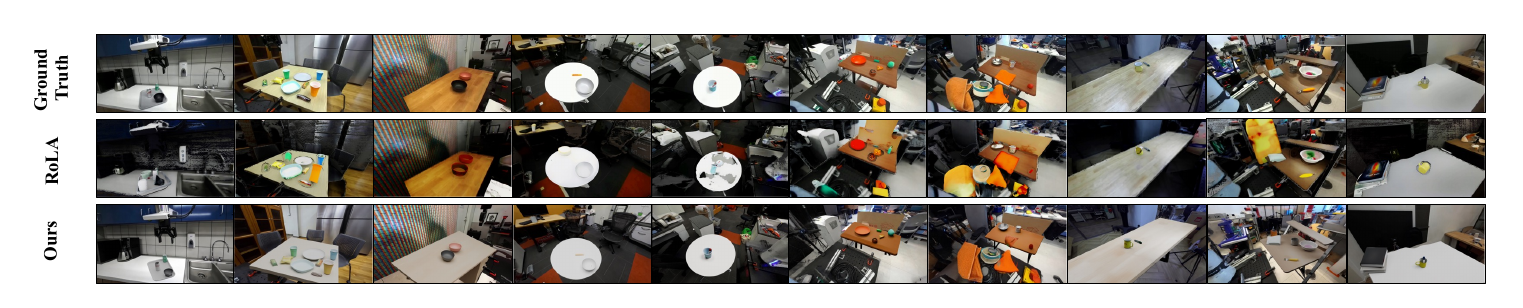}
\caption{\textbf{Quantitative and qualitative results.} Top: averages over 10 scenes. Bottom: visualizations under extrinsic camera settings.}
\label{fig:realism-visual}
\end{minipage}
\end{figure}

\paragraph{Visual Alignment.} We compare alignment to the input image with RoLA~\citep{rola2025} under the same single-frame setting, using each method's respective segmentation and reconstruction pipeline. Since \method reconstructs the full interaction region rather than preserving large background areas, PSNR and LPIPS~\citep{LPIPS} show only modest differences, while other metrics in Figure~\ref{fig:realism-visual} indicate better preservation of scene structure, color, and texture distribution. Metric definitions are in Appendix~\ref{app:visual-metrics}.

\paragraph{Simulation Readiness.}
We define a scene as \emph{simulation-ready} if it remains physically stable without severe floating or interpenetration after being loaded into a simulator. We import each reconstruction into Isaac Sim~\citep{isaacsim} and run the physics simulation for 300 physics steps. \method substantially improves stability metrics (Table~\ref{tab:realism-physics}), supporting downstream robot interaction and policy evaluation (\textbf{Q1}). Metric definitions are in Appendix~\ref{app:sim-metrics}.

We further ablate the refinement components on the same scene set. Physics-only refinement achieves the lowest falling rate but has a substantially higher collision rate than the full method (0.3561 vs.~0.0256). SDF-only refinement and removing graph constraints or support/contact losses also increase collision and pose-drift metrics. The full method achieves the lowest collision rate and all three pose-drift metrics, demonstrating the complementary roles of geometric constraints and physics settling in maintaining physical stability while preserving the recovered layout.

\begin{table}[htbp]
\centering
\small
\caption{Simulation stability and refinement ablations over 10
\dataset{} scenes after 300 Isaac Sim steps.
Lower is better; the best value in each column is bold.}
\label{tab:realism-physics}
\resizebox{\linewidth}{!}{%
\begin{tabular}{lccccc}
\toprule
Method
& Falling $\downarrow$
& Collision $\downarrow$
& \shortstack{Trans MSE\\(m$^2$) $\downarrow$}
& \shortstack{Mean disp.\\(m) $\downarrow$}
& Quat MSE $\downarrow$ \\
\midrule
SAM~3D~\citep{sam3d2025}
& 0.5640 & 0.3590 & 0.1079 & 0.3284 & 0.1560 \\
SAM~3D + FoundationPose~\citep{foundationpose2024}
& 0.5900 & 0.1538 & 0.1921 & 0.4383 & 0.1703 \\
RoLA~\citep{rola2025}
& 0.3810 & 0.3226 & 0.0736 & 0.2713 & 0.1093 \\
\midrule
\multicolumn{6}{l}{\textit{Refinement ablations of \method{}}} \\
Z-up only
& 0.3472 & 0.7111 & 0.1599 & 0.5210 & 0.1245 \\
w/o graph constraints
& 0.1200 & 0.4283 & 0.0101 & 0.1111 & 0.0209 \\
SDF only
& 0.3893 & 0.2663 & 0.0866 & 0.3334 & 0.1363 \\
Physics only
& \textbf{0.0700} & 0.3561 & 0.0115 & 0.1054 & 0.0286 \\
w/o support/contact losses
& 0.1150 & 0.2861 & 0.0139 & 0.1203 & 0.0414 \\
w/o refinement
& 0.4320 & 0.2982 & 0.0977 & 0.3126 & 0.1255 \\
\midrule
\textbf{\method{} (ours)}
& 0.1026 & \textbf{0.0256} & \textbf{0.0022}
& \textbf{0.0474} & \textbf{0.0178} \\
\bottomrule
\end{tabular}%
}
\end{table}

\subsection{Trajectory Replay}
\label{sec:exp-replay}

We evaluate replay of the original DROID trajectories on five scenes from the 10-scene subset in \S\ref{sec:exp-realism} where both methods yield stable and executable reconstructions.

This evaluation differs from demonstration-driven real-to-sim pipelines such as ReBot~\citep{fang2025rebot} and RialTo~\citep{rialto2024} in both its objective and the role of demonstration signals. ReBot uses gripper poses extracted from real-world trajectories to determine object placement in simulation for data augmentation. RialTo executes policies learned from real-world demonstrations in reconstructed environments to collect trajectories with privileged state information, which subsequently support reinforcement learning for policy robustification. In these pipelines, demonstration signals guide trajectory transfer or policy learning in simulation. 

\begin{wrapfigure}[17]{r}{0.5\linewidth}
\vspace{-0.8em}
\centering
\small
\setlength{\tabcolsep}{3pt}

\resizebox{\linewidth}{!}{%
\begin{tabular}{lcccccc}
\toprule
Method & Scene 1 & Scene 2 & Scene 3 & Scene 4 & Scene 5 & Sum \\
\midrule
\method & \cmark & \cmark & \cmark & \cmark & \cmark & 5/5 \\
RoLA~\citep{rola2025} & \cmark & \xmark & \cmark & \xmark & \xmark & 2/5 \\
\bottomrule
\end{tabular}%
}

\includegraphics[width=\linewidth]{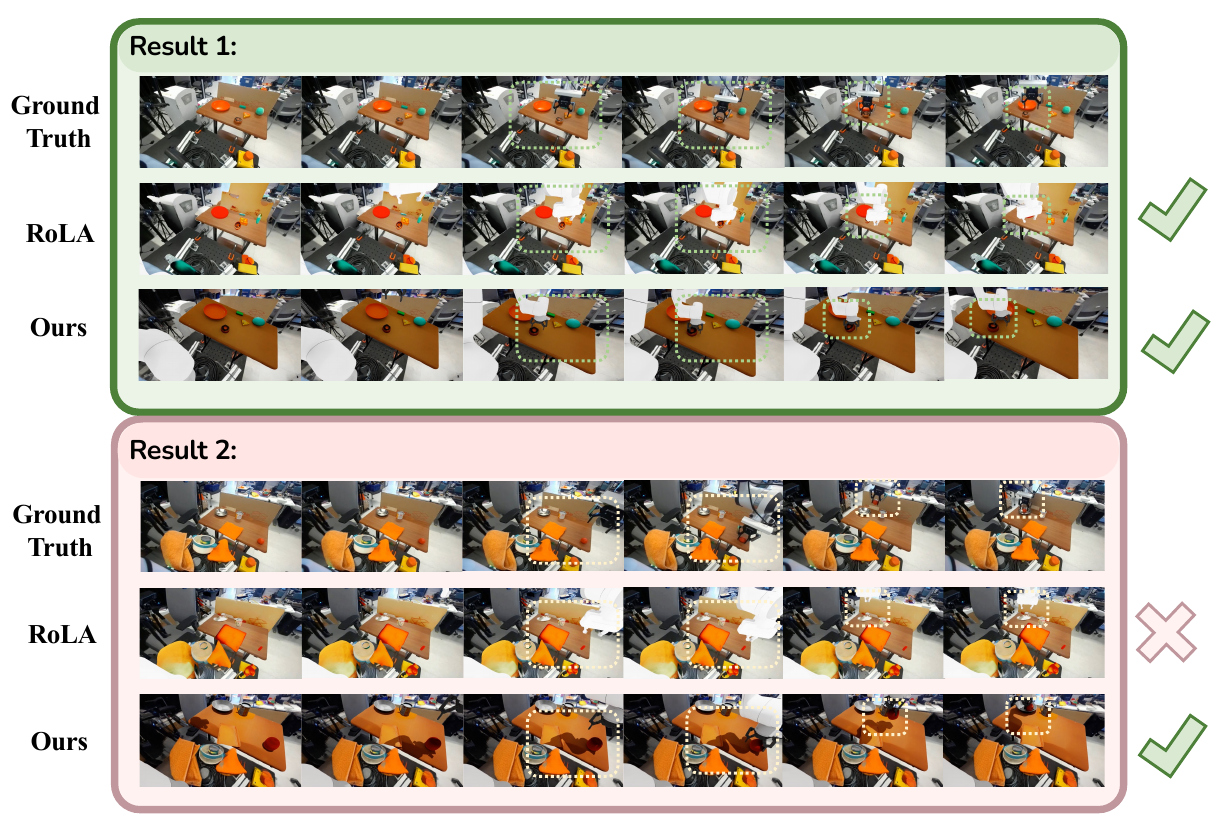}

\caption{\textbf{Replay examples visualization.} Full results are provided in the supplementary video.}
\label{fig:replay_exp}

\end{wrapfigure}

In contrast, our replay experiment evaluates the recovered scene itself (\textbf{Q2}): objects are instantiated according to the layout refined by \method, and recorded robot trajectories are replayed in the resulting environment. Here, the trajectories serve as probes of scene fidelity, while object placement is determined by the recovered layout. The experiment thus examines whether the recovered spatial configuration supports the corresponding robot-object interactions under replay, providing an interaction-based assessment of scene reconstruction beyond visual similarity.

A replay trial succeeds if the gripper grasps the intended object and moves it to the target without interpenetration or collisions. We compare with RoLA~\citep{rola2025}, a similar single-image scene recovery method that can support replay when action trajectories are given. Successful rollouts indicate that the recovered layout and robot base are accurate enough to reproduce key contact events in the original demonstration.

\par
\noindent

\subsection{Robot Data Generation}
\label{sec:exp-finetune}

We next explore whether demonstrations generated from \method scenes improve performance on user-defined tasks (\textbf{Q3}).

\begin{figure*}[htbp]
\centering

\begin{minipage}[t]{0.47\textwidth}
\vspace{0pt}
\centering
\includegraphics[width=\linewidth]{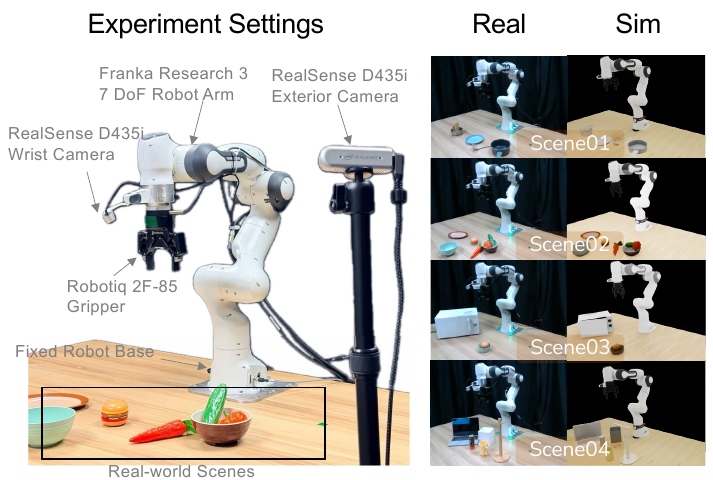}
\end{minipage}
\begin{minipage}[t]{0.44\textwidth}

\vspace{3pt}

\centering

\resizebox{0.80\linewidth}{!}{%
\begin{minipage}{\linewidth}
\vspace{14pt}
\centering
\footnotesize
\setlength{\tabcolsep}{3pt}
\renewcommand{\arraystretch}{0.98}
\begin{tabularx}{\linewidth}{c X}
\toprule
Task & Task description \\
\midrule
1.1 & Put the bread into the blue plate. \\
1.2 & Put the spoon in the pot. \\
\midrule
2.1 & Put the carrot on the desk into the bowl. \\
2.2 & Put the pumpkin into the blue bowl. \\
2.3 & Put the small hamburger in the plate. \\
\midrule
\textbf{3.1} & Open the white microwave. \\
\textbf{3.2} & Put the bowl into the microwave. \\
\midrule
4.1 & Put the can into the white bin. \\
4.2 & Close the laptop. \\
4.3 & Hang the mug on the wooden shelf. \\
\bottomrule
\end{tabularx}
\end{minipage}%
}
\end{minipage}

\caption{\textbf{Real-world evaluation setups and tasks.} Left: four real world scene setups. Right: task suite for each real-world scene. Tasks 3.1 and 3.2 are consecutive stages of a \textbf{long-horizon task}.}

\label{fig:real-tasks}

\end{figure*}

Across four real setups, we fine-tune $\pi_{0.5}$~\citep{pi05} and $\pi_0$~\citep{pi02024} under a real-only baseline and three streamed data-mixture settings over (real demonstrations, \method-generated demonstrations, simulation-augmented demonstrations): R1=$(0.2,0.4,0.4)$, R2=$(0.6,0.2,0.2)$, and R3=$(0,0.5,0.5)$. For each task, we collect 30 real demonstrations and evaluate each policy over 30 real-world trials. Detailed data synthesis and experiment settings are provided in Appendix~\ref{app:real-task}.

Ratio 2 yields the best average performance, improving success from 32.7\% to 41.7\% for $\pi_{0.5}$ and from 29.3\% to 42.7\% for $\pi_0$ as shown in Table~\ref{tab:finetune-main}. Ratio 3 still achieves nonzero real-world success without real demonstrations. These results show that \method scenes can generate useful task-specific data for real-world policy fine-tuning (\textbf{Q3}). Additional policy evaluations, augmentation baselines, and background ablations are provided in Appendix~\ref{app:real-task}.

\providecommand{\pmv}[2]{#1$\pm$#2}

\begin{table*}[htbp]
\centering
\caption{Real-world success rates (\%) over three 10-trial runs, reported as mean $\pm$ StdErr.}
\label{tab:finetune-main}
\footnotesize
\setlength{\tabcolsep}{1.2pt}
\renewcommand{\arraystretch}{1.03}
\begin{tabular*}{\textwidth}{@{\extracolsep{\fill}}ccccccccc@{}}
\toprule
\multirow{2}{*}{Task}
& \multicolumn{4}{c}{$\pi_{0.5}$}
& \multicolumn{4}{c}{$\pi_0$} \\
\cmidrule(lr){2-5}\cmidrule(lr){6-9}
& Real & R1 & R2 & R3
& Real & R1 & R2 & R3 \\
\midrule

1.1 & \pmv{40.0}{11.5} & \pmv{36.7}{8.8}  & \pmv{\textbf{63.3}}{6.7} & \pmv{20.0}{5.8}
    & \pmv{23.3}{3.3}  & \pmv{20.0}{5.8}  &  \pmv{\textbf{56.7}}{3.3} &  \pmv{10.0}{5.8}  \\

1.2 & \pmv{36.7}{3.3}  & \pmv{36.7}{6.7}  & \pmv{\textbf{50.0}}{5.8}  & \pmv{16.7}{3.3}

    & \pmv{33.3}{6.7}  & \pmv{36.7}{8.8}  & \pmv{\textbf{63.3}}{3.3}  & \pmv{13.3}{8.8} \\

\midrule

2.1 & \pmv{43.3}{3.3}  & \pmv{33.3}{3.3} & \pmv{\textbf{46.7}}{3.3} & \pmv{23.3}{6.7}
    & \pmv{43.3}{6.7}  & \pmv{33.3}{6.7} & \pmv{\textbf{50.0}}{5.8} & \pmv{16.7}{3.3} \\

2.2 & \pmv{26.7}{3.3}  & \pmv{36.7}{6.7} & \pmv{\textbf{43.3}}{8.8} & \pmv{13.3}{3.3}
    & \pmv{56.7}{3.3}  & \pmv{60.0}{11.5} & \pmv{\textbf{66.7}}{8.8} & \pmv{13.3}{3.3} \\

2.3 & \pmv{36.7}{3.3}  & \pmv{40.0}{0.0} & \pmv{\textbf{46.7}}{3.3} & \pmv{23.3}{3.3}
    & \pmv{16.7}{3.3}  & \pmv{13.3}{8.8} & \pmv{\textbf{20.0}}{5.8} & \pmv{6.7}{3.3} \\

\midrule

3.1 & \pmv{26.7}{5.8}  & \pmv{\textbf{36.7}}{3.3} & \pmv{30.0}{5.8} & \pmv{10.0}{5.8}
    & \pmv{30.0}{5.8}  & \pmv{46.7}{8.8} & \pmv{\textbf{56.7}}{8.8} & \pmv{33.3}{3.3} \\

3.2 & \pmv{6.7}{5.8}   & \pmv{6.7}{3.3}  & \pmv{3.3}{3.3}  & \pmv{0.0}{0.0}
    & \pmv{3.3}{3.3}   & \pmv{3.3}{3.3} & \pmv{\textbf{10.0}}{0.0} & \pmv{0.0}{0.0} \\

\midrule

4.1 & \pmv{36.7}{8.8}  & \pmv{30.0}{5.8} & \pmv{\textbf{36.7}}{3.3} & \pmv{16.7}{6.7}
    & \pmv{26.7}{3.3}  & \pmv{23.3}{6.7} & \pmv{23.3}{3.3} & \pmv{20.0}{5.8} \\

4.2 & \pmv{50.0}{5.8}  & \pmv{73.3}{6.7} & \pmv{\textbf{80.0}}{5.8} & \pmv{40.0}{5.8}
    & \pmv{43.3}{8.8}  & \pmv{53.3}{3.3} & \pmv{\textbf{56.7}}{8.8} & \pmv{30.0}{5.8} \\

4.3 & \pmv{23.3}{3.3}  & \pmv{\textbf{26.7}}{3.3} & \pmv{16.7}{6.7} & \pmv{10.0}{5.8}
    & \pmv{16.7}{3.3}  & \pmv{20.0}{0.0} & \pmv{\textbf{23.3}}{3.3} & \pmv{6.7}{6.7} \\

\midrule
Average & 32.7 & 35.7 & \textbf{41.7} & 17.3 & 29.3 & 31.0 & \textbf{42.7} & 15.0 \\
\bottomrule
\end{tabular*}
\end{table*}

\subsection{Randomization}
\label{sec:exp-robust}

Simulation enables controlled data augmentations that are costly to reproduce physically. We evaluate whether policies fine-tuned with \method-generated data retain performance under six real-world perturbations: object pose (\textbf{Obj.}, $\pm$10\,cm), background objects (\textbf{BG}), lighting (\textbf{Lt.}), table texture (\textbf{Tex.}), camera pose (\textbf{Cam.}), and robot initial state (\textbf{Arm}). We use $\pi_{0.5}$ performance under Ratio 2 from \S\ref{sec:exp-finetune}. Detailed settings are provided in Appendix~\ref{app:randomization}.

Table~\ref{fig:robust} reports the relative success-rate degradation from the original setting. Real-sim co-training reduces the average degradation from 13\% to 8\% across perturbation types, with clear gains under camera-pose and robot-initial-state shifts. These results show that data generated in \method scenes improves robustness retention under real-world perturbations (\textbf{Q4}).

\begin{table*}[htbp]

\centering
\caption{
\textbf{Robustness under perturbations.} Results are real-world success rates (\%) over 30 trials.
}
\label{fig:robust}
\vspace{-0.4em}

\resizebox{\textwidth}{!}{%
\begin{minipage}{\textwidth}
\centering

\begin{minipage}[htbp]{0.40\textwidth}
\vspace{3pt}
\centering
\includegraphics[width=1.06\linewidth]{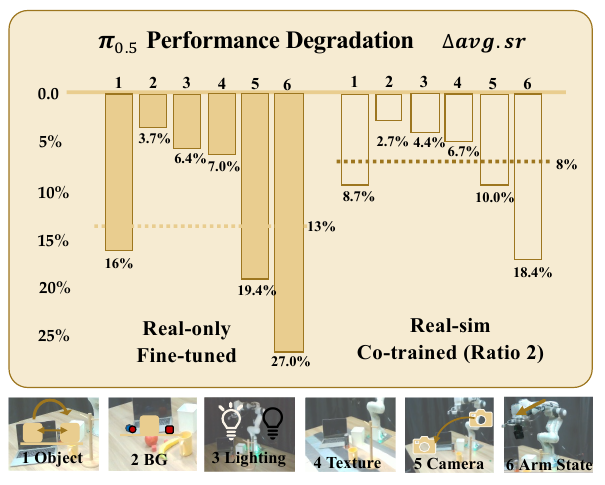}
\end{minipage}
\hspace{0.045\textwidth}
\begin{minipage}[htbp]{0.54\textwidth}
\vspace{0pt}
\centering
\tiny
\setlength{\tabcolsep}{1.4pt}
\renewcommand{\arraystretch}{0.99}
\resizebox{\linewidth}{!}{%
\begin{tabular}{c ccccccc ccccccc}
\toprule
\multirow{2}{*}{Task}
& \multicolumn{7}{c}{Real-only}
& \multicolumn{7}{c}{Mix Ratio 2} \\
\cmidrule(lr){2-8}
\cmidrule(lr){9-15}
& Orig. & Obj. & BG & Lt. & Tex. & Cam. & Arm
& Orig. & Obj. & BG & Lt. & Tex. & Cam. & Arm \\
\midrule

1.1
& 40.0 & 16.7 & 33.3 & 36.7 & 30.0 & 23.3 & 6.7
& 63.3 & 56.7 & 60.0 & 56.7 & 50.0 & 46.7 & 36.7 \\
1.2
& 36.7 & 20.0 & 33.3 & 30.0 & 26.7 & 16.7 & 3.3
& 50.0 & 43.3 & 46.7 & 46.7 & 43.3 & 40.0 & 26.7 \\

\midrule
2.1
& 43.3 & 26.7 & 36.7 & 40.0 & 33.3 & 13.3 & 10.0
& 46.7 & 26.7 & 46.7 & 43.3 & 36.7 & 43.3 & 20.0 \\
2.2
& 26.7 & 10.0 & 26.7 & 23.3 & 20.0 & 13.3 & 0.0
& 43.3 & 33.3 & 43.3 & 36.7 & 33.3 & 26.7 & 16.7 \\
2.3
& 36.7 & 23.3 & 30.0 & 33.3 & 26.7 & 10.0 & 6.7
& 46.7 & 36.7 & 36.7 & 40.0 & 30.0 & 36.7 & 23.3 \\

\midrule
3.1
& 26.7 & 13.3 & 23.3 & 20.0 & 26.7 & 16.7 & 10.0
& 30.0 & 26.7 & 33.3 & 26.7 & 30.0 & 23.3 & 20.0 \\
3.2
& 6.7 & 0.0 & 6.7 & 3.3 & 0.0 & 0.0 & 0.0
& 3.3 & 0.0 & 3.3 & 3.3 & 6.7 & 0.0 & 0.0 \\

\midrule
4.1
& 36.7 & 16.7 & 26.7 & 33.3 & 30.0 & 10.0 & 3.3
& 36.7 & 33.3 & 30.0 & 33.3 & 26.7 & 23.3 & 16.7 \\
4.2
& 50.0 & 40.0 & 53.3 & 36.7 & 43.3 & 23.3 & 13.3
& 80.0 & 66.7 & 76.7 & 73.3 & 76.7 & 70.0 & 63.3 \\
4.3
& 23.3 & 0.0 & 20.0 & 6.7 & 20.0 & 6.7 & 3.3
& 16.7 & 6.7 & 13.3 & 13.3 & 16.7 & 6.7 & 10.0 \\

\midrule
\textbf{Avg.}
& \textbf{32.7} & \textbf{16.7} & \textbf{29.0} & \textbf{26.3} & \textbf{25.7} & \textbf{13.3} & \textbf{5.66}
& \textbf{41.7} & \textbf{33.0} & \textbf{39.0} & \textbf{37.3} & \textbf{35.0} & \textbf{31.7} & \textbf{23.3} \\

\bottomrule
\end{tabular}%
}
\end{minipage}

\end{minipage}%
}

\vspace{-0.5em}
\end{table*}

\newpage
\subsection{Generative Evaluation Harness}
\label{sec:exp-correlation}

\begin{wrapfigure}[15]{r}{0.55\linewidth}
\vspace{-1.0em}
\centering
\small
\renewcommand{\arraystretch}{0.82}
\resizebox{0.98\linewidth}{!}{%
\begin{tabular}{lcccccccccc}
\toprule
Task & 1.1 & 1.2 & 2.1 & 2.2 & 2.3 & 3.1 & 3.2 & 4.1 & 4.2 & 4.3 \\
\midrule
Real SR & 40.0 & 36.7 & 43.3 & 26.7 & 36.7 & 26.7 & 6.7 & 36.7 & 50.0 & 23.3 \\
Sim SR  & 50.0 & 26.7 & 36.7 & 23.3 & 30.0 & 20.0  & 3.3   & 33.3   & 73.3 & 13.3   \\
\bottomrule
\end{tabular}%
}

\vspace{0.25em}
\includegraphics[width=\linewidth]{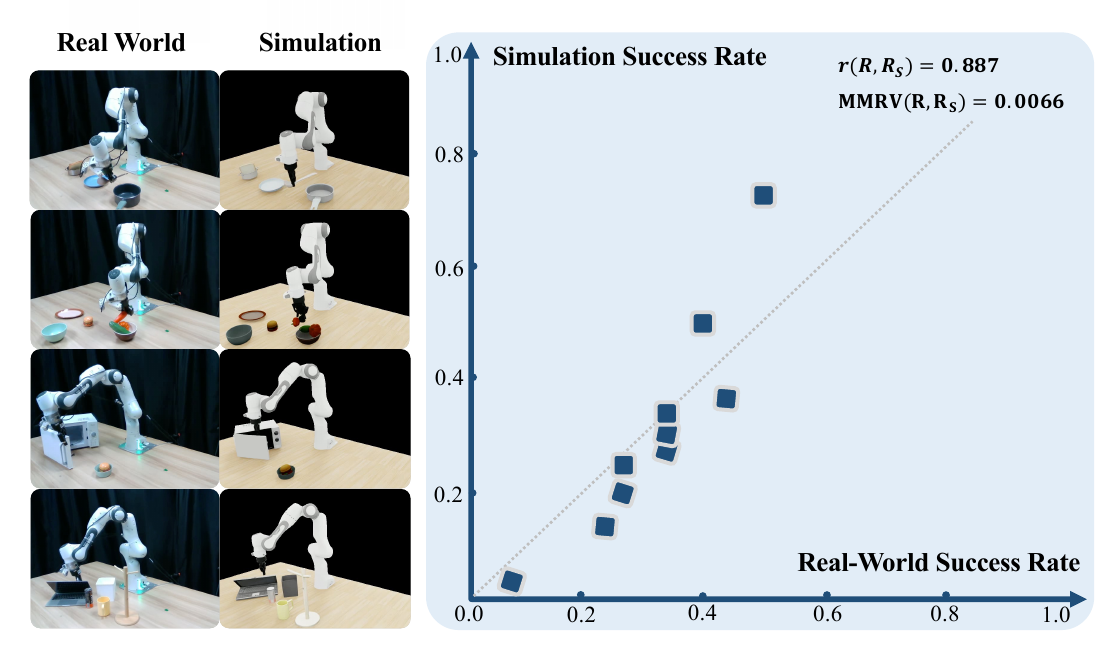}

\caption{\textbf{Sim-real correlation.} Real and simulated success rates (\%) of \textbf{real-only} fine-tuned $\pi_{0.5}$ policies.}
\label{fig:sim-real-correlation}
\vspace{-0.6em}
\end{wrapfigure}

\emph{Generative evaluation harness} (\textbf{Q5}) in our paper refers to treating synthetic environments as infrastructure for \textit{recursive self-improvement} of manipulation policies, with the aim of enabling more flexible and accessible policy evaluation than in the real world. This is challenging for manipulation because embodied evaluation depends on both visual realism and correct contact dynamics.

To assess whether \method scenes can serve as a generative evaluation harness, we run the \textbf{real-only} fine-tuned $\pi_{0.5}$ policies from \S\ref{sec:exp-finetune} in the corresponding generated scenes and compare real-world and simulation success rates. 

The resulting Pearson correlation ($r=0.887$) and MMRV ($\text{0.0066}$) indicate that \method captures sufficient sim-real correlation and task-relevant dynamics, which can provide a simulation proxy for real-world manipulation evaluation.

\section{Conclusion}
\label{sec:conclusion}

We presented \method, a single-image real-to-sim method that reconstructs scenes that can be
re-rendered, edited, and reused from new viewpoints. By separating interactive physical objects from visual context and refining object poses for simulation stability, \method produces scenes that are both visually faithful and physically stable. Our experimental results show that the recovered \method scenes can support faithful replay, generate useful robot data, and serve as a reliable evaluation harness for real-world manipulation. We view this as a beneficial trial toward treating real-world scene reconstruction not merely as building a visual digital twin, but as reusable infrastructure for embodied training and evaluation.

\section{Limitations}
\label{sec:limitations}

(1) \textbf{Input Quality.} \method reconstructs reusable scenes from single-RGB captures. Therefore, inputs of extreme low quality like severe occlusion, extreme lighting, and visually ambiguous materials can reduce reconstruction quality or reliability of generated demonstrations. 

(2) \textbf{Object and Physics Regime.} The current system focuses on rigid and articulated objects with well-defined support/contact structure. In this work, we do not target deformable, granular, or fluid materials, whose behavior is not well captured by rigid-body simulation models and is beyond the scope of this work.

(3) \textbf{Parameter Estimation Pipeline.} The system does not include a dedicated pipeline for automatic physical parameter estimation. Parameters such as friction and mass are inferred from VLM prior knowledge, while joint parameters for articulated objects are retrieved from objects of the similar type in the standard dataset.

(4) \textbf{Further Validation.} Since \method outputs a simulator-level scene rather than policy-specific data or labels, the same scene interface can in principle support different manipulation policies including video/world-model-based structures. Given our training and evaluation budget, we validate this interface on the settings and models in this paper and plan to leave broader framework validation to future work.



\clearpage

\bibliography{main}  

@InProceedings{Xiang_2020_SAPIEN,
author = {Xiang, Fanbo and Qin, Yuzhe and Mo, Kaichun and Xia, Yikuan and Zhu, Hao and Liu, Fangchen and Liu, Minghua and Jiang, Hanxiao and Yuan, Yifu and Wang, He and Yi, Li and Chang, Angel X. and Guibas, Leonidas J. and Su, Hao},
title = {{SAPIEN}: A SimulAted Part-based Interactive ENvironment},
booktitle = {The IEEE Conference on Computer Vision and Pattern Recognition (CVPR)},
month = {June},
year = {2020}}

@article{rialto2024,
  title   = {Reconciling Reality through Simulation: A Real-to-Sim-to-Real Approach for Robust Manipulation},
  author  = {Torne, Marcel and Simeonov, Anthony and Li, Zechu and Chan, April and Chen, Tao and Gupta, Abhishek and Agrawal, Pulkit},
  journal = {Robotics: Science and Systems (RSS)},
  year    = {2024}
}

@article{wang2026robovip,
  title={RoboVIP: Multi-View Video Generation with Visual Identity Prompting Augments Robot Manipulation},
  author={Wang, Boyang and Zhang, Haoran and Zhang, Shujie and Hao, Jinkun and Jia, Mingda and Lv, Qi and Mao, Yucheng and Lyu, Zhaoyang and Zeng, Jia and Xu, Xudong and others},
  journal={arXiv preprint arXiv:2601.05241},
  year={2026}
}

@article{robosplat,
  title={Novel Demonstration Generation with Gaussian Splatting Enables Robust One-Shot Manipulation},
  author={Yang, Sizhe and Yu, Wenye and Zeng, Jia and Lv, Jun and Ren, Kerui and Lu, Cewu and Lin, Dahua and Pang, Jiangmiao},
  journal={arXiv preprint arXiv:2504.13175},
  year={2025}
}

@inproceedings{grs2025,
  title     = {{GRS}: Generating Robotic Simulation Tasks from Real-World Images},
  author    = {Zook, Alex and Sun, Fan-Yun and Spjut, Josef and Blukis, Valts and Birchfield, Stan and Tremblay, Jonathan},
  booktitle = {Proceedings of the IEEE/CVF Conference on Computer Vision and Pattern Recognition Workshops},
  pages     = {594--603},
  year      = {2025}
}

@article{fang2025rebot,
  title={ReBot: Scaling Robot Learning with Real-to-Sim-to-Real Robotic Video Synthesis},
  author={Fang, Yu and Yang, Yue and Zhu, Xinghao and Zheng, Kaiyuan and Bertasius, Gedas and Szafir, Daniel and Ding, Mingyu},
  journal={arXiv preprint arXiv:2503.14526},
  year={2025}
}

@inproceedings{acdc2024,
  title={Automated Creation of Digital Cousins for Robust Policy Learning},
  author={Tianyuan Dai and Josiah Wong and Yunfan Jiang and Chen Wang and Cem Gokmen and Ruohan Zhang and Jiajun Wu and Li Fei-Fei},
  booktitle={Conference on Robot Learning (CoRL)},
  year={2024}
}

@inproceedings{yang2024physcene,
          title={PhyScene: Physically Interactable 3D Scene Synthesis for Embodied AI},
          author={Yang, Yandan and Jia, Baoxiong and Zhi, Peiyuan and Huang, Siyuan},
          booktitle={Proceedings of Conference on Computer Vision and Pattern Recognition (CVPR)},
          year={2024}
}

@inproceedings{yang2025sceneweaver,
          title={SceneWeaver: All-in-One 3D Scene Synthesis with an Extensible and Self-Reflective Agent},
          author={Yang, Yandan and Jia, Baoxiong and Zhang, Shujie and Huang, Siyuan},
          booktitle = {Advances in Neural Information Processing Systems (NeurIPS)},
          year={2025}
}

@inproceedings{yu2025metascenes,
  title={METASCENES: Towards Automated Replica Creation for Real-world 3D Scans},
  author={Yu, Huangyue and Jia, Baoxiong and Chen, Yixin and Yang, Yandan and Li, Puhao and Su, Rongpeng and Li, Jiaxin and Li, Qing and Liang, Wei and Zhu Song-Chun and Liu, Tengyu and Huang, Siyuan},
  booktitle={Conference on Computer Vision and Pattern Recognition(CVPR)},
  year={2025}
}

@inproceedings{rola2025,
  title={Robot learning from any images},
  author={Zhao, Siheng and Mao, Jiageng and Chow, Wei and Shangguan, Zeyu and Shi, Tianheng and Xue, Rong and Zheng, Yuxi and Weng, Yijia and You, Yang and Seita, Daniel and others},
  booktitle={Conference on Robot Learning},
  pages={4226--4245},
  year={2025},
  organization={PMLR}
}

@misc{polaris2025,
      title={PolaRiS: Scalable Real-to-Sim Evaluations for Generalist Robot Policies}, 
      author={Arhan Jain and Mingtong Zhang and Kanav Arora and William Chen and Marcel Torne and Muhammad Zubair Irshad and Sergey Zakharov and Yue Wang and Sergey Levine and Chelsea Finn and Wei-Chiu Ma and Dhruv Shah and Abhishek Gupta and Karl Pertsch},
      year={2025},
      eprint={2512.16881},
      archivePrefix={arXiv},
      primaryClass={cs.RO},
      url={https://arxiv.org/abs/2512.16881}, 
}

@inproceedings{re3sim2025,
  title={RE$^3$SIM: Generating High-Fidelity Simulation Data via 3D-Photorealistic Real-to-Sim for Robotic Manipulation},
  author={Han, Xiaoshen and Yu, Junqiu and Liu, Minghuan and Chen, Yilun and Lyu, Xiaoyang and Tian, Yang and Wang, Bolun and Zhang, Weinan and Zhang, Weinan and Pang, Jiangmiao},
  booktitle={IEEE International Conference on Robotics and Automation (ICRA)},
  year={2026}
}

@misc{splatsim2024,
            title={SplatSim: Zero-Shot Sim2Real Transfer of RGB Manipulation Policies Using Gaussian Splatting}, 
            author={Mohammad Nomaan Qureshi and Sparsh Garg and Francisco Yandun and David Held and George Kantor and Abhishesh Silwal},
            year={2024},
            eprint={2409.10161},
            archivePrefix={arXiv},
            primaryClass={cs.RO},
            url={https://arxiv.org/abs/2409.10161}, 
}

@article{twinaligner2025,
author    = {Hongwei Fan and Hang Dai and Jiyao Zhang and Jinzhou Li and Qiyang Yan and Yujie Zhao and Mingju Gao and Jinghang Wu and Hao Tang and Hao Dong},
title     = {TwinAligner: Visual-Dynamic Alignment Empowers Physics-aware Real2Sim2Real for Robotic Manipulation},
year={2025},
eprint={2512.19390},
archivePrefix={arXiv},
primaryClass={cs.RO},
url={https://arxiv.org/abs/2512.19390},
}

@article{real2edit2real2025,
      title={Real2Edit2Real: Generating Robotic Demonstrations via a 3D Control Interface}, 
      author={Yujie Zhao and Hongwei Fan and Di Chen and Shengcong Chen and Liliang Chen and Xiaoqi Li and Guanghui Ren and Hao Dong},
      year={2025},
      eprint={2512.19402},
      archivePrefix={arXiv},
      primaryClass={cs.RO},
      url={https://arxiv.org/abs/2512.19402}, 
}

@article{scalable_real2sim2025,
    author        = {Pfaff, Nicholas and Fu, Evelyn and Binagia, Jeremy and Isola, Phillip and Tedrake, Russ},
    title         = {Scalable Real2Sim: Physics-Aware Asset Generation Via Robotic Pick-and-Place Setups},
    year          = {2025},
    eprint        = {2503.00370},
    archivePrefix = {arXiv},
    primaryClass  = {cs.RO},
    url           = {https://arxiv.org/abs/2503.00370}, 
  }

@article{roboengine2025,
  title={RoboEngine: Plug-and-Play Robot Data Augmentation with Semantic Robot Segmentation and Background Generation},
  author={Yuan, Chengbo and Joshi, Suraj and Zhu, Shaoting and Su, Hang and Zhao, Hang and Gao, Yang},
  journal={arXiv preprint arXiv:2503.18738},
  year={2025}
}

@misc{robogs2024,
  title={Robo-GS: A Physics Consistent Spatial-Temporal Model for Robotic Arm with Hybrid Representation}, 
  author={Haozhe Lou and Yurong Liu and Yike Pan and Yiran Geng and Jianteng Chen and Wenlong Ma and Chenglong Li and Lin Wang and Hengzhen Feng and Lu Shi and Liyi Luo and Yongliang Shi},
  year={2024},
  eprint={2408.14873},
  archivePrefix={arXiv},
  primaryClass={cs.RO},
  url={https://arxiv.org/abs/2408.14873}, 
}

@article{mv_sam3d2026,
  title={MV-SAM3D: Adaptive Multi-View Fusion for Layout-Aware 3D Generation},
  author={Li, Baicheng and Wu, Dong and Li, Jun and Zhou, Shunkai and Zeng, Zecui and Li, Lusong and Zha, Hongbin},
  journal={arXiv preprint arXiv:2603.11633},
  year={2026}
}

@misc{real2render2real2025,
      title={Real2Render2Real: Scaling Robot Data Without Dynamics Simulation or Robot Hardware}, 
      author={Justin Yu and Letian Fu and Huang Huang and Karim El-Refai and Rares Andrei Ambrus and Richard Cheng and Muhammad Zubair Irshad and Ken Goldberg},
      year={2025},
      eprint={2505.09601},
      archivePrefix={arXiv},
      primaryClass={cs.RO},
      url={https://arxiv.org/abs/2505.09601}, 
}

@inproceedings{robocasa2024,
  title={RoboCasa: Large-Scale Simulation of Everyday Tasks for Generalist Robots},
  author={Soroush Nasiriany and Abhiram Maddukuri and Lance Zhang and Adeet Parikh and Aaron Lo and Abhishek Joshi and Ajay Mandlekar and Yuke Zhu},
  booktitle={Robotics: Science and Systems (RSS)},
  year={2024}
}

@article{simplerenv2024,
          title={Evaluating Real-World Robot Manipulation Policies in Simulation},
          author={Xuanlin Li and Kyle Hsu and Jiayuan Gu and Karl Pertsch and Oier Mees and Homer Rich Walke and Chuyuan Fu and Ishikaa Lunawat and Isabel Sieh and Sean Kirmani and Sergey Levine and Jiajun Wu and Chelsea Finn and Hao Su and Quan Vuong and Ted Xiao},
          journal = {arXiv preprint arXiv:2405.05941},
          year={2024},
}

@misc{oxe2023,
  title     = {Open {X-E}mbodiment: Robotic Learning Datasets and {RT-X} Models},
  author    = {Open X-Embodiment Collaboration and Abby O'Neill and Abdul Rehman and et al.},
  howpublished  = {\url{https://arxiv.org/abs/2310.08864}},
  year = {2023},
}

@article{droid2024,
  title={Droid: A large-scale in-the-wild robot manipulation dataset},
  author={Khazatsky, Alexander and Pertsch, Karl and Nair, Suraj and Balakrishna, Ashwin and Dasari, Sudeep and Karamcheti, Siddharth and Nasiriany, Soroush and Srirama, Mohan Kumar and Chen, Lawrence Yunliang and Ellis, Kirsty and others},
  journal={arXiv preprint arXiv:2403.12945},
  year={2024}
}

@inproceedings{bridgev2,
    title={BridgeData V2: A Dataset for Robot Learning at Scale},
    author={Walke, Homer and Black, Kevin and Lee, Abraham and Kim, Moo Jin and Du, Max and Zheng, Chongyi and Zhao, Tony and Hansen-Estruch, Philippe and Vuong, Quan and He, Andre and Myers, Vivek and Fang, Kuan and Finn, Chelsea and Levine, Sergey},
    booktitle={Conference on Robot Learning (CoRL)},
    year={2023}
}

@article{pi02024,
  title={{$\pi$0}: A vision-language-action flow model for general robot control},
  author={Black, Kevin and Brown, Noah and Driess, Danny and Esmail, Adnan and Equi, Michael and Finn, Chelsea and others},
  journal={arXiv preprint arXiv:2410.24164},
  year={2024}
}

@article{pi05,
  title={{$\pi$0.5}: A vision-language-action model with open-world generalization},
  author={Physical Intelligence},
  journal={arXiv preprint arXiv:2504.16054},
  year={2025}
}

@article{openvla2024,
    title={OpenVLA: An Open-Source Vision-Language-Action Model},
    author={{Moo Jin} Kim and Karl Pertsch and Siddharth Karamcheti and Ted Xiao and Ashwin Balakrishna and Suraj Nair and Rafael Rafailov and Ethan Foster and Grace Lam and Pannag Sanketi and Quan Vuong and Thomas Kollar and Benjamin Burchfiel and Russ Tedrake and Dorsa Sadigh and Sergey Levine and Percy Liang and Chelsea Finn},
    journal = {arXiv preprint arXiv:2406.09246},
    year={2024}
}

@inproceedings{mimicgen2023,
    title={MimicGen: A Data Generation System for Scalable Robot Learning using Human Demonstrations},
    author={Mandlekar, Ajay and Nasiriany, Soroush and Wen, Bowen and Akinola, Iretiayo and Narang, Yashraj and Fan, Linxi and Zhu, Yuke and Fox, Dieter},
    booktitle={7th Annual Conference on Robot Learning},
    year={2023}
}

@inproceedings{dexmimicgen2024,
      title     = {DexMimicGen: Automated Data Generation for Bimanual Dexterous Manipulation via Imitation Learning},
      author    = {Jiang, Zhenyu and Xie, Yuqi and Lin, Kevin and Xu, Zhenjia and Wan, Weikang and Mandlekar, Ajay and Fan, Linxi and Zhu, Yuke},
      booktitle = {2025 IEEE International Conference on Robotics and Automation (ICRA)},
      year      = {2025}
}

@inproceedings{vggt2025,
  title={VGGT: Visual Geometry Grounded Transformer},
  author={Wang, Jianyuan and Chen, Minghao and Karaev, Nikita and Vedaldi, Andrea and Rupprecht, Christian and Novotny, David},
  booktitle={Proceedings of the IEEE/CVF Conference on Computer Vision and Pattern Recognition},
  year={2025}
}

@article{liu2023grounding,
  title={Grounding dino: Marrying dino with grounded pre-training for open-set object detection},
  author={Liu, Shilong and Zeng, Zhaoyang and Ren, Tianhe and Li, Feng and Zhang, Hao and Yang, Jie and Li, Chunyuan and Yang, Jianwei and Su, Hang and Zhu, Jun and others},
  journal={arXiv preprint arXiv:2303.05499},
  year={2023}
}

@article{xiao2023florence, title={Florence-2: Advancing a unified representation for a variety of vision tasks}, author={Xiao, Bin and Wu, Haiping and Xu, Weijian and Dai, Xiyang and Hu, Houdong and Lu, Yumao and Zeng, Michael and Liu, Ce and Yuan, Lu}, journal={arXiv preprint arXiv:2311.06242}, year={2023} }

@InProceedings{foundationpose2024,
author        = {Bowen Wen and Wei Yang and Jan Kautz and Stan Birchfield},
title         = {{FoundationPose}: Unified 6D Pose Estimation and Tracking of Novel Objects},
booktitle     = {CVPR},
year          = {2024},
}

@inproceedings{depth_pro2024,
  author     = {Aleksei Bochkovskii and Ama\"{e}l Delaunoy and Hugo Germain and Marcel Santos and
               Yichao Zhou and Stephan R. Richter and Vladlen Koltun},
  title      = {Depth Pro: Sharp Monocular Metric Depth in Less Than a Second},
  booktitle  = {International Conference on Learning Representations},
  year       = {2025},
  url        = {https://arxiv.org/abs/2410.02073},
}

@article{LPIPS,
  title   = {The Unreasonable Effectiveness of Deep Features as a Perceptual Metric},
  author  = {Zhang, Richard and Isola, Phillip and Efros, Alexei A. and Shechtman, Eli and Wang, Oliver},
  journal = {arXiv preprint arXiv:1801.03924},
  year    = {2018}
}

@article{sam3d2025,
      title={SAM 3D: 3Dfy Anything in Images}, 
      author={SAM 3D Team and Xingyu Chen and Fu-Jen Chu and Pierre Gleize and Kevin J Liang and Alexander Sax and Hao Tang and Weiyao Wang and Michelle Guo and Thibaut Hardin and Xiang Li and Aohan Lin and Jiawei Liu and Ziqi Ma and Anushka Sagar and Bowen Song and Xiaodong Wang and Jianing Yang and Bowen Zhang and Piotr Dollár and Georgia Gkioxari and Matt Feiszli and Jitendra Malik},
      year={2025},
      eprint={2511.16624},
      archivePrefix={arXiv},
      primaryClass={cs.CV},
      url={https://arxiv.org/abs/2511.16624}, 
}

@article{tian2025interndata,
  title={InternData-A1: Pioneering high-fidelity synthetic data for pre-training generalist policy},
  author={Tian, Yang and Yang, Yuyin and Xie, Yiman and Cai, Zetao and Shi, Xu and Gao, Ning and Liu, Hangxu and Jiang, Xuekun and Qiu, Zherui and Yuan, Feng and others},
  journal={arXiv preprint arXiv:2511.16651},
  year={2025}
}

@misc{sam3_2025,
      title={SAM 3: Segment Anything with Concepts},
      author={Nicolas Carion and Laura Gustafson and Yuan-Ting Hu and Shoubhik Debnath and Ronghang Hu and Didac Suris and Chaitanya Ryali and Kalyan Vasudev Alwala and Haitham Khedr and Andrew Huang and Jie Lei and Tengyu Ma and Baishan Guo and Arpit Kalla and Markus Marks and Joseph Greer and Meng Wang and Peize Sun and Roman Rädle and Triantafyllos Afouras and Effrosyni Mavroudi and Katherine Xu and Tsung-Han Wu and Yu Zhou and Liliane Momeni and Rishi Hazra and Shuangrui Ding and Sagar Vaze and Francois Porcher and Feng Li and Siyuan Li and Aishwarya Kamath and Ho Kei Cheng and Piotr Dollár and Nikhila Ravi and Kate Saenko and Pengchuan Zhang and Christoph Feichtenhofer},
      year={2025},
      eprint={2511.16719},
      archivePrefix={arXiv},
      primaryClass={cs.CV},
      url={https://arxiv.org/abs/2511.16719},
}

@article{trellis2025,
    title   = {Structured 3D Latents for Scalable and Versatile 3D Generation},
    author  = {Xiang, Jianfeng and Lv, Zelong and Xu, Sicheng and Deng, Yu and Wang, Ruicheng and Zhang, Bowen and Chen, Dong and Tong, Xin and Yang, Jiaolong},
    journal = {arXiv preprint arXiv:2412.01506},
    year    = {2024}
}

@article{zhou2018open3d,
    author    = {Qian-Yi Zhou and Jaesik Park and Vladlen Koltun},
    title     = {{Open3D}: {A} Modern Library for {3D} Data Processing},
    journal   = {arXiv:1801.09847},
    year      = {2018},
}

@misc{hunyuan3d2025,
    title={Hunyuan3D 2.0: Scaling Diffusion Models for High Resolution Textured 3D Assets Generation},
    author={Tencent Hunyuan3D Team},
    year={2025},
    eprint={2501.12202},
    archivePrefix={arXiv},
    primaryClass={cs.CV}
}

@article{achiam2023gpt,
  title={Gpt-4 technical report},
  author={Achiam, Josh and Adler, Steven and Agarwal, Sandhini and Ahmad, Lama and Akkaya, Ilge and Aleman, Florencia Leoni and Almeida, Diogo and Altenschmidt, Janko and Altman, Sam and Anadkat, Shyamal and others},
  journal={arXiv preprint arXiv:2303.08774},
  year={2023}
}

@article{yang2023set,
  title={Set-of-mark prompting unleashes extraordinary visual grounding in gpt-4v},
  author={Yang, Jianwei and Zhang, Hao and Li, Feng and Zou, Xueyan and Li, Chunyuan and Gao, Jianfeng},
  journal={arXiv preprint arXiv:2310.11441},
  year={2023}
}

@article{yao2025cast,
  title={Cast: Component-aligned 3d scene reconstruction from an rgb image},
  author={Yao, Kaixin and Zhang, Longwen and Yan, Xinhao and Zeng, Yan and Zhang, Qixuan and Xu, Lan and Yang, Wei and Gu, Jiayuan and Yu, Jingyi},
  journal={ACM Transactions on Graphics (TOG)},
  volume={44},
  number={4},
  pages={1--19},
  year={2025},
  publisher={ACM New York, NY, USA}
}

@misc{mesatask2025,
  title={MesaTask: Towards Task-Driven Tabletop Scene Generation via 3D Spatial Reasoning}, 
  author={Hao, Jinkun and Liang, Naifu and Luo, Zhen and Xu, Xudong and Zhong, Weipeng and Yi, Ran and Jin, Yichen and Lyu, Zhaoyang and Zheng, Feng and Ma, Lizhuang and Pang, Jiangmiao},
  journal={arXiv preprint arXiv:2509.22281},
  year={2025}
}

@misc{internscenes2025,
      title={InternScenes: A Large-scale Simulatable Indoor Scene Dataset with Realistic Layouts}, 
      author={Weipeng Zhong and Peizhou Cao and Yichen Jin and Li Luo and Wenzhe Cai and Jingli Lin and Hanqing Wang and Zhaoyang Lyu and Tai Wang and Bo Dai and Xudong Xu and Jiangmiao Pang},
      year={2026},
      eprint={2509.10813},
      archivePrefix={arXiv},
      primaryClass={cs.CV},
      url={https://arxiv.org/abs/2509.10813}, 
}

@article{vhacd_collision,
  title={Volumetric hierarchical approximate convex decomposition},
  author={Mamou, Khaled and Lengyel, E and Peters, A},
  journal={Game engine gems},
  volume={3},
  pages={141--158},
  year={2016},
  publisher={AK Peters}
}

@software{isaacsim,
author = {{NVIDIA}},
license = {Apache-2.0},
title = {{Isaac Sim}},
url = {https://github.com/isaac-sim/IsaacSim},
version = {6.0.1}
}

@inproceedings{maniskill2,
  title={ManiSkill2: A Unified Benchmark for Generalizable Manipulation Skills},
  author={Gu, Jiayuan and Xiang, Fanbo and Li, Xuanlin and Ling, Zhan and Liu, Xiqiaing and Mu, Tongzhou and Tang, Yihe and Tao, Stone and Wei, Xinyue and Yao, Yunchao and Yuan, Xiaodi and Xie, Pengwei and Huang, Zhiao and Chen, Rui and Su, Hao},
  booktitle={International Conference on Learning Representations},
  year={2023}
}

@inproceedings{rt1,
    title={RT-1: Robotics Transformer for Real-World Control at Scale},
    author={Anthony	Brohan and  Noah Brown and  Justice Carbajal and  Yevgen Chebotar and  Joseph Dabis and  Chelsea Finn and  Keerthana Gopalakrishnan and  Karol Hausman and  Alex Herzog and others},
    booktitle={arXiv preprint arXiv:2212.06817},
    year={2022}
}

@inproceedings{octo,
    title={Octo: An Open-Source Generalist Robot Policy},
    author = {{Octo Model Team} and Dibya Ghosh and Homer Walke and Karl Pertsch and Kevin Black and Oier Mees and Sudeep Dasari and Joey Hejna and Charles Xu and Jianlan Luo and Tobias Kreiman and {You Liang} Tan and Lawrence Yunliang Chen and Pannag Sanketi and Quan Vuong and Ted Xiao and Dorsa Sadigh and Chelsea Finn and Sergey Levine},
    booktitle = {Proceedings of Robotics: Science and Systems},
    address  = {Delft, Netherlands},
    year = {2024},
}

@article{chen2025robotwin,
  title={Robotwin 2.0: A scalable data generator and benchmark with strong domain randomization for robust bimanual robotic manipulation},
  author={Chen, Tianxing and Chen, Zanxin and Chen, Baijun and Cai, Zijian and Liu, Yibin and Li, Zixuan and Liang, Qiwei and Lin, Xianliang and Ge, Yiheng and Gu, Zhenyu and others},
  journal={arXiv preprint arXiv:2506.18088},
  year={2025}
}

@article{liu2023libero,
  title={LIBERO: Benchmarking Knowledge Transfer for Lifelong Robot Learning},
  author={Liu, Bo and Zhu, Yifeng and Gao, Chongkai and Feng, Yihao and Liu, Qiang and Zhu, Yuke and Stone, Peter},
  journal={arXiv preprint arXiv:2306.03310},
  year={2023}
}

@inproceedings{robocasa365,
  title={RoboCasa365: A Large-Scale Simulation Framework for Training and Benchmarking Generalist Robots},
  author={Soroush Nasiriany and Sepehr Nasiriany and Abhiram Maddukuri and Yuke Zhu},
  booktitle={International Conference on Learning Representations (ICLR)},
  year={2026}
}

@article{anygrasp2023,
  title={AnyGrasp: Robust and Efficient Grasp Perception in Spatial and Temporal Domains},
  author = {Fang, Hao-Shu and Wang, Chenxi and Fang, Hongjie and Gou, Minghao and Liu, Jirong and Yan, Hengxu and Liu, Wenhai and Xie, Yichen and Lu, Cewu},
  journal={IEEE Transactions on Robotics (T-RO)},
  year={2023}
}

@misc{curobo2026,
      title={cuRoboV2: Dynamics-Aware Motion Generation with Depth-Fused Distance Fields for High-DoF Robots},
      author={Balakumar Sundaralingam and Adithyavairavan Murali and Stan Birchfield},
      year={2026},
      eprint={2603.05493},
      archivePrefix={arXiv},
      primaryClass={cs.RO}
}

@misc{choi2026scalingsimtorealreinforcementlearning,
      title={Scaling Sim-to-Real Reinforcement Learning for Robot VLAs with Generative 3D Worlds}, 
      author={Andrew Choi and Xinjie Wang and Zhizhong Su and Wei Xu},
      year={2026},
      eprint={2603.18532},
      archivePrefix={arXiv},
      primaryClass={cs.RO},
      url={https://arxiv.org/abs/2603.18532}, 
}

@inproceedings{wang2026vggtomega,
  title     = {{VGGT-$\Omega$}},
  author    = {Jianyuan Wang and Minghao Chen and Shangzhan Zhang and Nikita Karaev and Johannes Sch{\"o}nberger and Patrick Labatut and Piotr Bojanowski and David Novotny and Andrea Vedaldi and Christian Rupprecht},
  booktitle = {Proceedings of the IEEE/CVF Conference on Computer Vision and Pattern Recognition (CVPR)},
  year      = {2026}
}

@article{wang2025tabletopgen,
  title={TabletopGen: Instance-Level Interactive 3D Tabletop Scene Generation from Text or Single Image},
  author={Wang, Ziqian and He, Yonghao and Yang, Licheng and Zou, Wei and Ma, Hongxuan and Liu, Liu and Sui, Wei and Guo, Yuxin and Su, Hu},
  journal={arXiv preprint arXiv:2512.01204},
  year={2025}
}

@article{behavior1k,
    title   = {BEHAVIOR-1K: A Human-Centered, Embodied AI Benchmark with 1,000 Everyday Activities and Realistic Simulation},
    author  = {Chengshu Li and Ruohan Zhang and Josiah Wong and Cem Gokmen and Sanjana Srivastava and Roberto Martín-Martín and Chen Wang and Gabrael Levine and Wensi Ai and Benjamin Martinez and Hang Yin and Michael Lingelbach and Minjune Hwang and Ayano Hiranaka and Sujay Garlanka and Arman Aydin and Sharon Lee and Jiankai Sun and Mona Anvari and Manasi Sharma and Dhruva Bansal and Samuel Hunter and Kyu-Young Kim and Alan Lou and Caleb R Matthews and Ivan Villa-Renteria and Jerry Huayang Tang and Claire Tang and Fei Xia and Yunzhu Li and Silvio Savarese and Hyowon Gweon and C. Karen Liu and Jiajun Wu and Li Fei-Fei},
    journal = {arXiv preprint arXiv:2403.09227},
    year    = {2024}
}

@misc{marble2025,
  title        = {Marble},
  author       = {{World Labs}},
  year         = {2026},
  howpublished = {\url{https://docs.worldlabs.ai/}}
}

@misc{nanobanana2025,
  title        = {Gemini 2.5 Flash Image (Nano Banana)},
  author       = {{Google AI for Developers}},
  year         = {2026},
  howpublished = {\url{https://ai.google.dev/gemini-api/docs/models/gemini-2.5-flash-image}},
  note         = {Model documentation. Accessed: 2026-05-16}
}

@inproceedings{infinigen2023infinite,
  title={Infinite Photorealistic Worlds Using Procedural Generation},
  author={Raistrick, Alexander and Lipson, Lahav and Ma, Zeyu and Mei, Lingjie and Wang, Mingzhe and Zuo, Yiming and Kayan, Karhan and Wen, Hongyu and Han, Beining and Wang, Yihan and Newell, Alejandro and Law, Hei and Goyal, Ankit and Yang, Kaiyu and Deng, Jia},
  booktitle={Proceedings of the IEEE/CVF Conference on Computer Vision and Pattern Recognition},
  pages={12630--12641},
  year={2023}
}

@misc{p3sam2025,
      title={P3-SAM: Native 3D Part Segmentation}, 
      author={Changfeng Ma and Yang Li and Xinhao Yan and Jiachen Xu and Yunhan Yang and Chunshi Wang and Zibo Zhao and Yanwen Guo and Zhuo Chen and Chunchao Guo},
      year={2025},
      eprint={2509.06784},
      archivePrefix={arXiv},
      primaryClass={cs.CV},
      url={https://arxiv.org/abs/2509.06784}, 
}

@article{robogen2024,
  title={Robogen: Towards unleashing infinite data for automated robot learning via generative simulation},
  author={Wang, Yufei and Xian, Zhou and Chen, Feng and Wang, Tsun-Hsuan and Wang, Yian and Fragkiadaki, Katerina and Erickson, Zackory and Held, David and Gan, Chuang},
  journal={arXiv preprint arXiv:2311.01455},
  year={2023}
}

\clearpage
\appendix

\section{Experiment}

\subsection{Metrics}
\subsubsection{Visual-alignment metrics}
\label{app:visual-metrics}

We evaluate visual alignment by comparing each method's rendered RGB image $\hat I$ to the original DROID frame $I$, resizing rendered images to match the ground-truth resolution with bicubic interpolation. We report six complementary metrics: pixel fidelity, structure, perceptual similarity, local geometry, color distribution, and texture statistics.

\paragraph{Pixel, structure, and perceptual metrics.}
PSNR is computed from RGB MSE after normalizing to $[0,1]$:
\[
\mathrm{MSE} = \frac{1}{3|\Omega|} \sum_{\mathbf u\in\Omega}\sum_{c\in\{r,g,b\}} (I_c(\mathbf u)-\hat I_c(\mathbf u))^2,\quad
\mathrm{PSNR} = 20\log_{10}\frac{1}{\sqrt{\mathrm{MSE}}}.
\]
SSIM uses \texttt{skimage.metrics.structural\_similarity} on uint8 RGB images; LPIPS-Alex uses the official AlexNet implementation on $[-1,1]$ scaled RGB. Higher PSNR/SSIM and lower LPIPS-Alex indicate better agreement.

\paragraph{Local-feature, color, and texture metrics.}
SIFT match ratio is computed on grayscale $640\times360$ images with up to 800 keypoints per image, using brute-force $\ell_2$ KNN matching and Lowe's 0.75 ratio test:
\[
\mathrm{SIFT\text{-}MR} = \frac{N_{\mathrm{good}}}{\min(N_I,N_{\hat I})+\epsilon}.
\]
RGB-Wasserstein compares 64-bin histograms per channel via $W_1$:
\[
\mathrm{RGB\text{-}W} = \frac{1}{3}\sum_c W_1(p_c,\hat p_c),\quad\text{lower is better.}
\]
Gabor-$\ell_1$ uses $K=24$ filters (8 orientations × 3 wavelengths) on $256\times144$ grayscale images; filter response energies are $\ell_2$-normalized and compared by
\[
\mathrm{Gabor\text{-}\ell_1} = \frac{1}{K}\Big\|\frac{\mathbf e(I)}{\|\mathbf e(I)\|_2+\epsilon}-\frac{\mathbf e(\hat I)}{\|\mathbf e(\hat I)\|_2+\epsilon}\Big\|_1.
\]
Lower Gabor-$\ell_1$ indicates closer texture-frequency agreement.

\subsubsection{Simulation-Ready Metrics}
\label{app:sim-metrics}

\begin{figure}[htbp]
\centering
\includegraphics[width=0.95\linewidth]{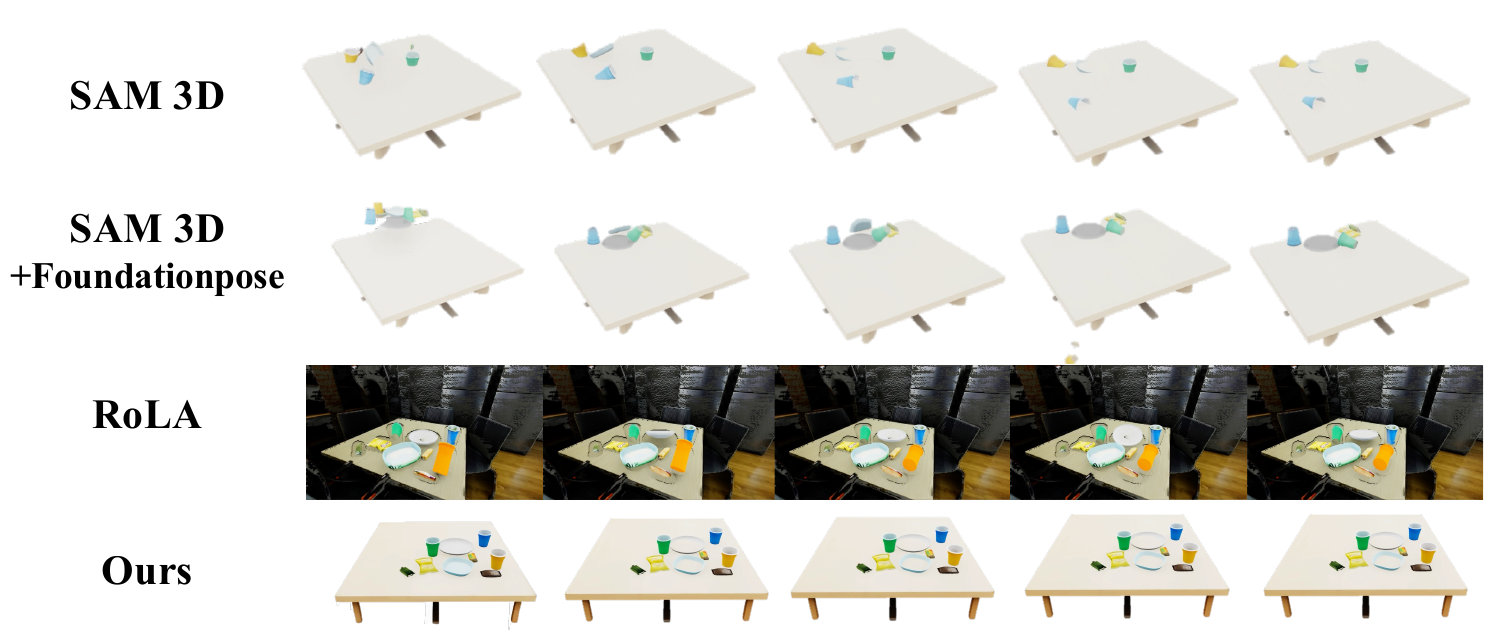}
\vspace{-0.5em}
\caption{\textbf{Simulation-readiness under gravity.}
We compare reconstructed scenes after loading them into IsaacLab and rolling out physics under gravity. \method -refined scene remains physically stable while preserving the recovered object arrangement.}
\label{fig:simulation-readiness}
\end{figure}

We measure whether a reconstructed scene is simulation-ready by first aligning it to a gravity-consistent coordinate frame. Specifically, we fit the dominant support plane with RANSAC and rotate the scene so that the fitted plane normal aligns with the simulator's $+z$ up axis, i.e., gravity acts along $-z$. We then load all object USD assets into IsaacLab with the transformed recovered poses and collision geometry, and step $T=300$ physics steps with gravity. Object root poses are sampled every 10 steps.

Let object $i$ have poses $\mathbf{x}_{i,t}=(\mathbf{p}_{i,t},\mathbf{q}_{i,t})$, with world-space center
\[
\mathbf{c}_{i,t} = \mathbf{p}_{i,t} + R(\mathbf{q}_{i,t})\,\mathbf{c}^{\mathrm{local}}_i,
\quad
\mathbf{c}^{\mathrm{local}}_i = R(\mathbf{q}_{i,0})^\top (\mathbf{c}_{i,0}^{\mathrm{world}} - \mathbf{p}_{i,0}).
\]
Translation drift and per-object metrics are
\[
\Delta \mathbf{c}_i = \mathbf{c}_{i,T}-\mathbf{c}_{i,0},\quad
\mathrm{TransMSE}_i = \frac{1}{3}\|\Delta \mathbf{c}_i\|_2^2,\quad
\mathrm{Disp}_i = \|\Delta \mathbf{c}_i\|_2.
\]
Rotation drift is computed after sign-correcting quaternions:
\[
\tilde{\mathbf{q}}_{i,T} =
\begin{cases}
\mathbf{q}_{i,T}, & \mathbf{q}_{i,0}^{\top}\mathbf{q}_{i,T} \ge 0,\\
-\mathbf{q}_{i,T}, & \mathbf{q}_{i,0}^{\top}\mathbf{q}_{i,T} < 0
\end{cases},\quad
\mathrm{QuatMSE}_i = \frac{1}{4}\|\tilde{\mathbf{q}}_{i,T}-\mathbf{q}_{i,0}\|_2^2.
\]

Falling is flagged if the tilt angle of the object’s local up exceeds $45^\circ$:
\[
\phi_i = \cos^{-1}\frac{\mathbf{u}_{i,0}^{\top}\mathbf{u}_{i,T}}{\|\mathbf{u}_{i,0}\|_2\|\mathbf{u}_{i,T}\|_2},\quad
\mathbf{u}_{i,t}=R(\mathbf{q}_{i,t})(0,0,1)^\top.
\]
Collision/pop-out failures are detected if consecutive center displacements exceed 0.05\,m or upward jumps exceed 0.03\,m. Scene-level metrics report fractions of falling/collision objects, mean Trans MSE, mean center displacement, and mean quaternion MSE. These jointly measure physical stability and simulation readiness.

\paragraph{Evaluation Horizon and Repeatability.}
The physics timestep is 0.01\,s, so the 300-step evaluation horizon corresponds to 3\,s of simulated time. The object-motion curves show that the main settling dynamics have largely decayed before this horizon for the compared methods. We additionally repeat the stability evaluation 10 times for each of the 10 scenes. \method{} has a mean settling time of 0.12\,s, with a mean within-scene standard deviation of 0.02\,s.

\begin{figure}[htbp]
  \vspace{-0.5em}
  \centering
  \includegraphics[width=\linewidth]{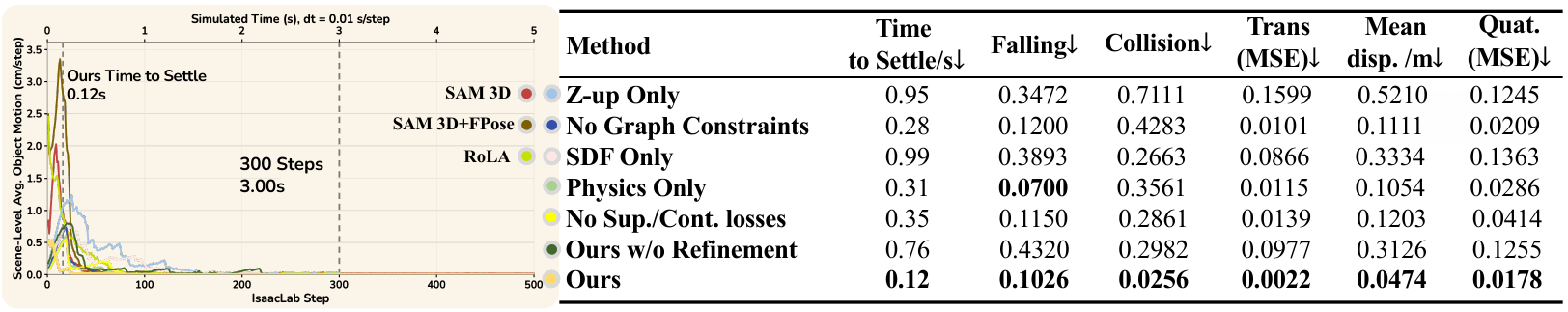}
  \vspace{-1em}
  \caption{Sim-ready ablations and settling dynamics. \textbf{Zoom in for best view}.%
  \label{fig:sim-ready}}
  \vspace{-0.8em}
\end{figure}

\subsection{Trajectory Replay}

We evaluate whether a recovered static scene can support open-loop replay of the original DROID trajectory. For each sampled scene, we instantiate the recovered objects, collision assets, gravity-aligned scene frame, and robot base in IsaacLab, then replay the recorded gripper trajectory in the recovered robot/world frame.

\begin{figure}[htbp]
\centering
\includegraphics[width=\linewidth]{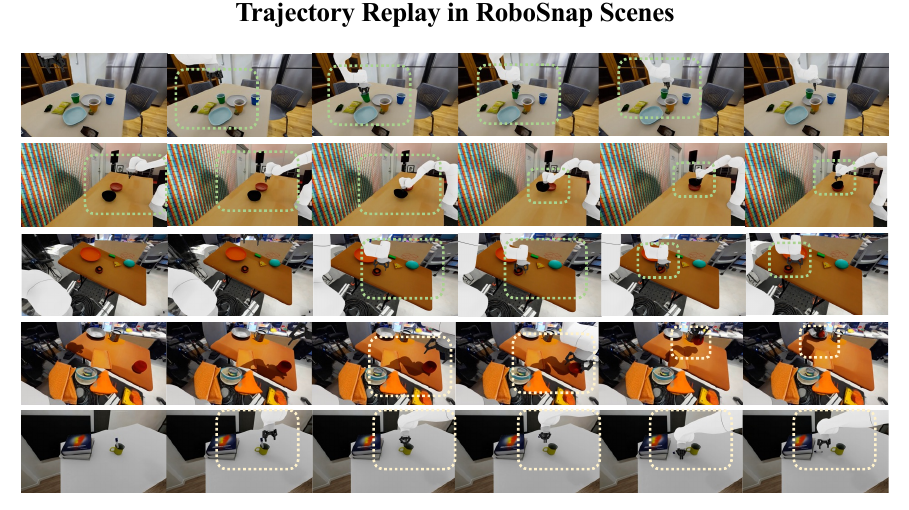}
\vspace{-0.5em}
\caption{\textbf{Replay of real DROID trajectories.}
We replay the same recorded gripper trajectory in scenes reconstructed by RoLA and \method. Dashed boxes mark key contact regions.}
\label{fig:replay}
\end{figure}

The trajectory is used only for evaluation: we do not optimize object poses from replay outcomes or use privileged simulation rollouts to repair failures. A trial succeeds if the gripper grasps the intended object and moves it to the target region without severe collision, interpenetration, or contact failure.

This test differs from simulation data generation with real-world replay and demonstration-driven real-to-sim pipelines. Rather than generating trajectories in simulation and transferring them to the real world, or using demonstrations to fit object placements, we ask whether an existing real trajectory becomes executable once the static scene, object layout, and robot base are accurately recovered. Thus, replay success measures active interaction fidelity of the reconstructed scene and suggests a low-cost path for augmenting real robot logs through simulation replay, perturbation, and re-rendering.

The replay experiment evaluates whether a reconstructed static scene can support the original contact sequence from a real demonstration. For each scene, we instantiate reconstructed objects with their recovered poses and collision geometry, place the robot using the robot-base transform in the dataset. Therefore, replay success primarily reflects the accuracy of scene recovery.

\subsection{Real Scene Tasks}
\label{app:real-task}

\begin{figure*}[htbp]
\centering

\begin{minipage}[t]{0.47\textwidth}
\vspace{0pt}
\centering
\includegraphics[width=0.93\linewidth]{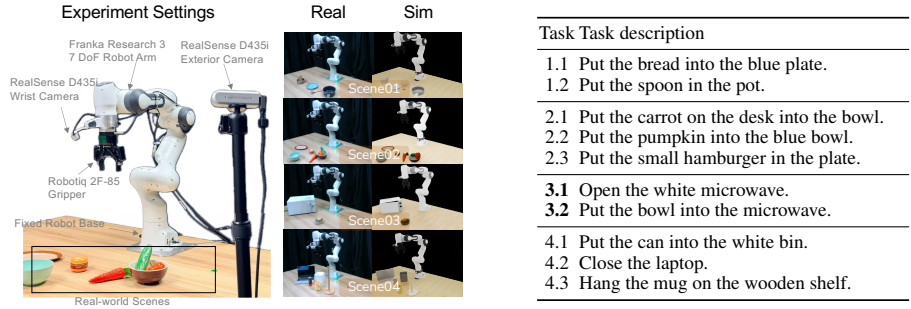}
\end{minipage}
\begin{minipage}[t]{0.44\textwidth}

\vspace{3pt}

\centering

\resizebox{0.80\linewidth}{!}{%
\begin{minipage}{\linewidth}
\vspace{6pt}
\centering
\footnotesize
\setlength{\tabcolsep}{1pt}
\renewcommand{\arraystretch}{0.98}
\begin{tabularx}{\linewidth}{c X}
\toprule
Task & Task description \\
\midrule
1.1 & Put the bread into the blue plate. \\
1.2 & Put the spoon in the pot. \\
\midrule
2.1 & Put the carrot on the desk into the bowl. \\
2.2 & Put the pumpkin into the blue bowl. \\
2.3 & Put the small hamburger in the plate. \\
\midrule
\textbf{3.1} & Open the white microwave. \\
\textbf{3.2} & Put the bowl into the microwave. \\
\midrule
4.1 & Put the can into the white bin. \\
4.2 & Close the laptop. \\
4.3 & Hang the mug on the wooden shelf. \\
\bottomrule
\end{tabularx}
\end{minipage}%
}
\end{minipage}

\caption{\textbf{Real-world evaluation setups and tasks.} Left: four real world scene setups. Right: task suite for each real-world scene. Tasks 3.1 and 3.2 are consecutive stages of a \textbf{long-horizon task}.}

\label{fig:real-tasks}
\end{figure*}

\newpage

\subsubsection{Training Details}

\begin{wrapfigure}{r}{0.45\linewidth}
\vspace{-1.2em}
\centering
\includegraphics[width=\linewidth]{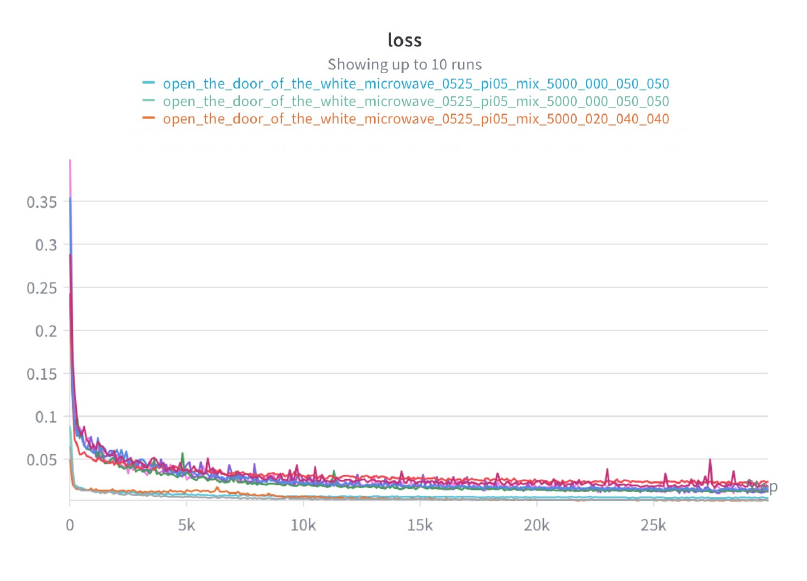}
\vspace{-1.0em}
\caption{\textbf{Training diagnostic.}
We plot the fine-tuning loss, i.e., the conditional flow-matching objective averaged over action dimensions, action horizon, and mini-batch samples. The loss largely plateaus near 15k steps; we therefore use the same 15k-step checkpoint for all settings to avoid unfair checkpoint selection.}
\label{fig:openpi-training-loss}
\vspace{-2.0em}
\end{wrapfigure}

For each task, we convert both real and generated trajectories into the RLDS format used by the OpenPI framework. The real-only baseline is initialized from the corresponding pretrained checkpoint and fine-tuned using 30 real demonstrations. For mixed-data settings, we additionally stream from two task-specific synthetic pools generated in the corresponding \method scene: simulation-generated demonstrations and simulation-augmented demonstrations. Each synthetic source contains 2,000 demonstrations per task. The simulation-augmented pool is evenly balanced across six perturbation types: manipulated-object displacement, background objects, lighting, camera pose, robot-arm initial state, and table texture.

All settings use the same optimization budget: batch size 16, a cosine learning-rate schedule with 100 warmup steps and peak learning rate $2\times10^{-5}$. We evaluate all methods at the fixed 15k-step checkpoint according to Figure ~\ref{fig:openpi-training-loss}. 

Data mixtures are implemented by streaming from the real, simulation-generated, and simulation-augmented RLDS sources with the specified weights. Thus, R1--R3 only change the expected source composition of consumed mini-batches, while keeping the number of gradient updates, batch size, model initialization, action space, and evaluation protocol fixed. We use joint-position actions with action horizon 16 for $\pi_{0.5}$, and use the corresponding OpenPI fine-tuning configuration for $\pi_0$.

For both $\pi_0$ and $\pi_{0.5}$, the logged training loss is the OpenPI action-generation objective rather than a task-success metric. Given an observation $o$ and a ground-truth action chunk $a$, the model samples Gaussian noise $\epsilon$ and a time $t$, constructs
\[
x_t = t\epsilon + (1-t)a,
\qquad
u_t = \epsilon - a,
\]
and predicts the velocity target $u_t$ from $(o,x_t,t)$. The reported scalar loss is
\[
\mathcal{L}
=
\mathbb{E}
\left[
\left\|
v_\theta(o,x_t,t)-u_t
\right\|_2^2
\right],
\]
averaged over action dimensions, action horizon, and mini-batch samples. Therefore, the loss curves are used as optimization diagnostics.

\subsubsection{Synthetic Data Generation}

We modify the InternDataEngine~\citep{tian2025interndata} to improve the sim-to-real performance. Specifically, we filter out irrational IK-based pick poses that are unlikely to succeed on real hardware. We also generate more regularized place poses by keeping them aligned with the preceding poses, which improves task success rate, accelerates trajectory generation, and further enhances transferability from simulation to the real world. These optimizations enable large-scale data generation, reaching approximately 10,000 task-specific trajectories per day on 8 RTX 4090 GPUs.

For each real setup, we export the recovered \method scene into a task-generation configuration. The configuration bridges the recovered scene and the data-generation engine: it specifies the scene assets, robot and camera setup, task instruction, target objects, skill template, and output format, which makes the task definition and trajectory synthesis flexible enough. A simplified example is shown below.

\begin{yamlbox}
task:
name: <scene-task-id>
scene:
asset_root: <robosnap-scene-root>
arena_file: <arena-config>

robot:
name: franka
config: <robot-config>
base_pose: <recovered-base-pose>

cameras:
wrist: <wrist-camera-config>
exterior: <exterior-camera-config>

objects:
target:
asset: <recovered-object-usd>
pose: <recovered-world-pose>
receptacle:
asset: <recovered-object-usd>
pose: <recovered-world-pose>

generation:
instruction: <task-language-instruction>
target_object: target
support_region: <task-specific-region>
skill_sequence:
- <pick/place/close/open-skill-template>

output:
task_dir: <output-directory>
save:
- rgb
- robot_state
- actions
- camera_params
- metadata
\end{yamlbox}

InternDataEngine loads this configuration to instantiate the recovered scene in Isaac Sim~\citep{isaacsim}, attach the specified robot and cameras, and execute the skill sequence under the language instruction. The generation pipeline samples task-relevant object poses or grasp candidates, produces 6-DoF end-effector waypoints, converts them into collision-aware robot actions, and saves successful rollouts as task-specific simulated demonstrations.

\begin{figure}[htbp]
\centering
\includegraphics[width=\linewidth]{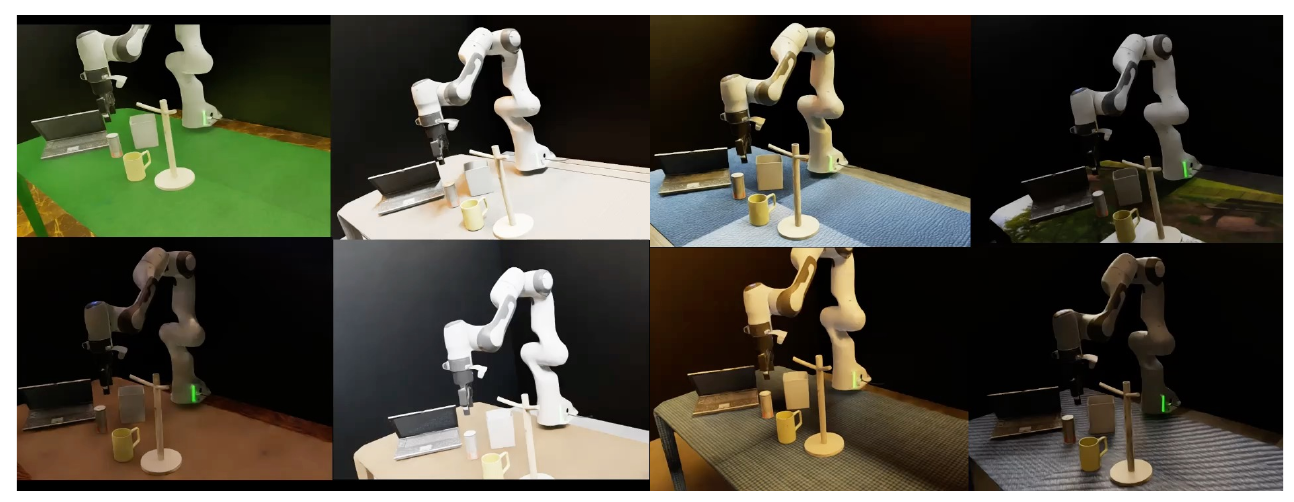}
\vspace{-0.5em}
\caption{\textbf{Synthetic trajectories augmentation.} Object poses, viewpoints, lighting, and appearance are uniformly perturbed to produce diverse task-consistent simulated demonstrations.}
\label{fig:generated-data}
\end{figure}

\subsubsection{Real World Evaluation}

We evaluate all policies on the physical Franka Research 3 setup with a Robotiq 2F-85 gripper. The robot model used in simulation is the standard Franka--Robotiq asset rather than a reconstructed or Gaussian-splat asset. A trial is counted as successful only if the task reaches the completion state specified by the language instruction and remains stable at the end of execution. For non-long-horizon tasks, partial completion is counted as failure; for the long-horizon microwave task, the two stages are evaluated separately as Tasks 3.1 and 3.2.

Before each trial, objects are reset to the prescribed initial positions and poses. Non-long-horizon trials are executed for at most 2,000 inference steps, while long-horizon trials are executed for at most 5,000 steps. If the task is not completed within the step budget, or if execution is aborted due to unsafe motion, severe collision, or invalid robot state, the trial is marked as failure. Each setting is evaluated with three batches of 10 trials, and we compute summary statistics over the three 10-trial runs.

\subsubsection{Additional Policy Evaluation}
\label{app:additional-policies}

We additionally evaluate Diffusion Policy (DP) and GR00T N1.6
on two representative real-world tasks: placing a spoon in a pot
(Task~1.2) and closing a laptop (Task~4.2). For each policy,
we compare real-only training with the R2 setting, which samples
real, simulation-generated, and simulation-augmented demonstrations
with weights $(0.6, 0.2, 0.2)$.

\begin{table}[htbp]
\centering
\small
\caption{Real-world success rates (\%) of additional policies
on two representative tasks.}
\label{tab:additional-policies}
\begin{tabular}{lcccc}
\toprule
& \multicolumn{2}{c}{Task 1.2}
& \multicolumn{2}{c}{Task 4.2} \\
\cmidrule(lr){2-3}
\cmidrule(lr){4-5}
Policy & Real-only & R2 & Real-only & R2 \\
\midrule
DP
& 30.0 & \textbf{46.7} & 53.3 & \textbf{76.7} \\
GR00T N1.6
& 43.3 & \textbf{53.3} & 63.3 & \textbf{86.7} \\
\bottomrule
\end{tabular}
\end{table}

As shown in Table~\ref{tab:additional-policies}, R2 improves
real-world success rates for both policies on both tasks.
These results provide supplementary evidence that
\method{}-generated demonstrations benefit additional policy
architectures in the evaluated settings.

\subsubsection{Augmentation Baseline and Background Ablation}
\label{app:augmentation-ablation}

Using $\pi_{0.5}$, we compare the full \method{} R2 setting
with a RoboEngine augmentation baseline and an R2 variant
without the background visual layer (w/o BG).
We evaluate Task~1.2 and Task~4.2, following the same
data-mixture recipe across the augmented-data settings.

\begin{table}[htbp]
\centering
\small
\caption{Real-world success rates (\%) of $\pi_{0.5}$
under different training-data settings on two representative tasks.}
\label{tab:augmentation-ablation}
\begin{tabular}{lcccc}
\toprule
Task & Real-only & RoboEngine (R2)
& R2 w/o BG & \method{} (R2) \\
\midrule
1.2 & 36.7 & 36.7 & 43.3 & \textbf{50.0} \\
4.2 & 50.0 & 53.3 & 66.7 & \textbf{80.0} \\
\bottomrule
\end{tabular}%
\end{table}

The full R2 setting achieves the highest success rates on
both evaluated tasks (Table~\ref{tab:augmentation-ablation}).
Removing the background reduces performance, suggesting
that preserving visual context contributes to the effectiveness
of the generated training data in these settings.

\subsection{Real-World Robustness Perturbations}
\label{app:randomization}

Starting from the original evaluation setup (Orig.), we introduce six controlled real-world perturbations to test policy robustness while keeping the task instruction and target object identity fixed.

In the object-pose condition (Obj.), the manipulated object is translated by approximately 10\,cm along a fixed direction before rollout. 

\begin{wrapfigure}{r}{0.4\linewidth}
\centering
\includegraphics[width=0.8\linewidth]{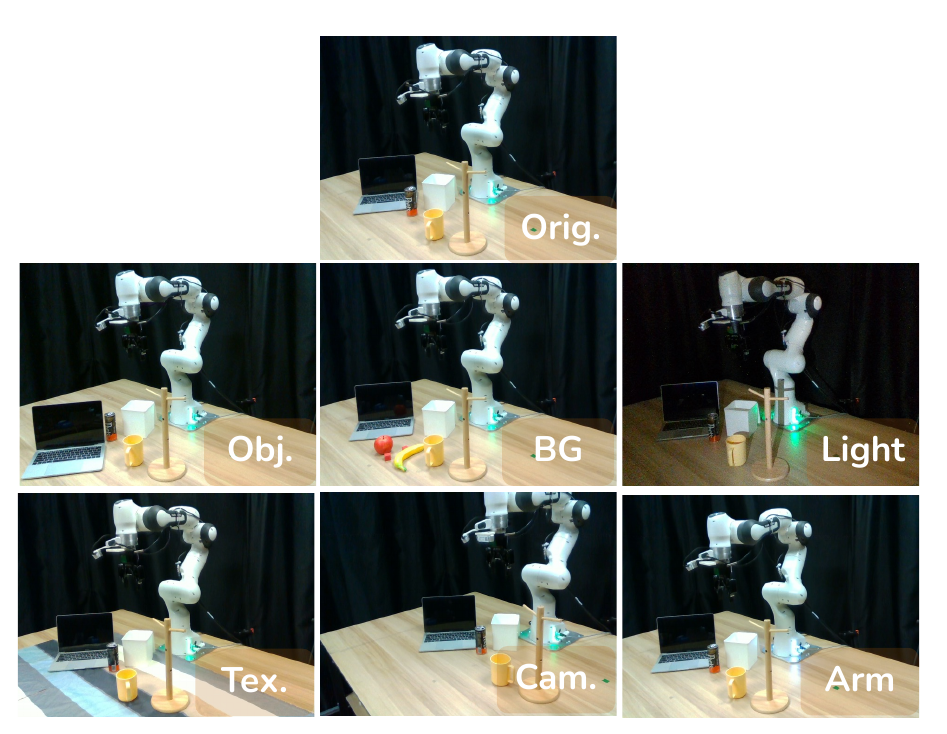}
\caption{\textbf{Real-world perturbations for robustness evaluation.}}
\label{fig:real-perturbations}
\vspace{-1.0em}
\end{wrapfigure}

In the background condition (BG), additional distractor objects, including cubes and fruit, are placed on the table. In the lighting condition (Light), the room light is turned off to change image illumination. In the texture condition (Tex.), the table appearance is changed by covering it with the same tablecloth across trials. In the camera condition (Cam.), the camera is moved to another fixed viewpoint. In the robot initial-state condition (Arm.), the end-effector is first driven to a fixed alternative starting pose before executing the policy. Each perturbation is applied independently, so performance changes can be attributed to a specific source of distribution shift.

\subsection{Generative Evaluation Harness}

\begin{wrapfigure}[10]{r}{0.55\textwidth}
  \centering
  \vspace{-1.0em}
  \includegraphics[width=\linewidth]{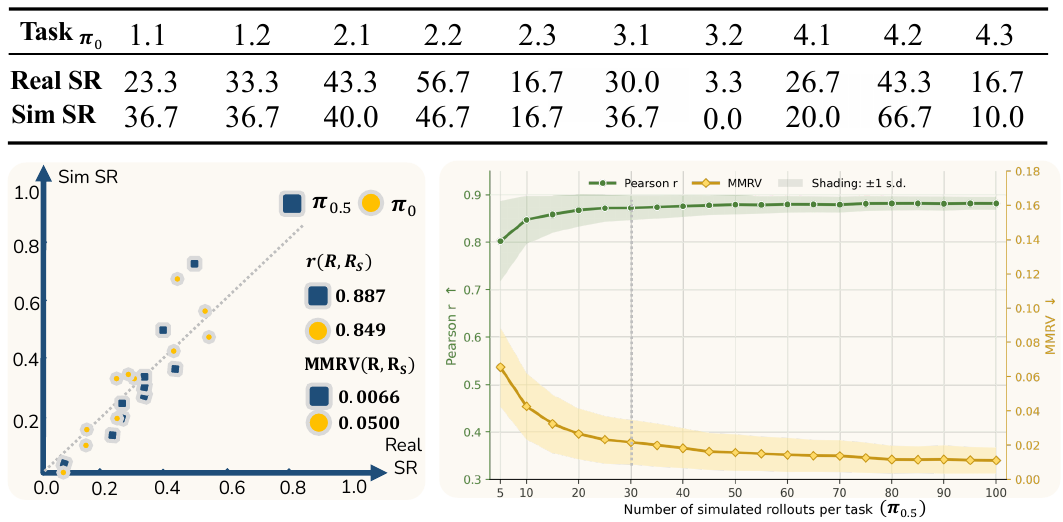}
  \caption{Additional sim--real alignment evidence.}
  \label{fig:section3_figure5}
\end{wrapfigure}

\paragraph{Additional Policy Alignment.}
We additionally evaluate the real-only fine-tuned $\pi_0$
policy in the corresponding real and simulated environments
on the same 10 tasks. The resulting Pearson correlation is
0.849 and MMRV is 0.0500, providing additional evidence for
sim--real alignment beyond the $\pi_{0.5}$ policy
(Figure~\ref{fig:section3_figure5}).

\paragraph{Sensitivity to Evaluation Budget.}
For $\pi_{0.5}$, we vary the number of simulated rollouts
per task and compute the alignment metrics over 20 batches
at each budget. Correlation becomes relatively stable at
around 30 rollouts per task as its variability decreases,
supporting this evaluation budget in the tested settings.

\paragraph{Behavior-Space Diagnostics for Generative Evaluation.}

\begin{wrapfigure}{r}{0.50\linewidth}
\vspace{-1.2em}
\centering
\includegraphics[width=\linewidth]{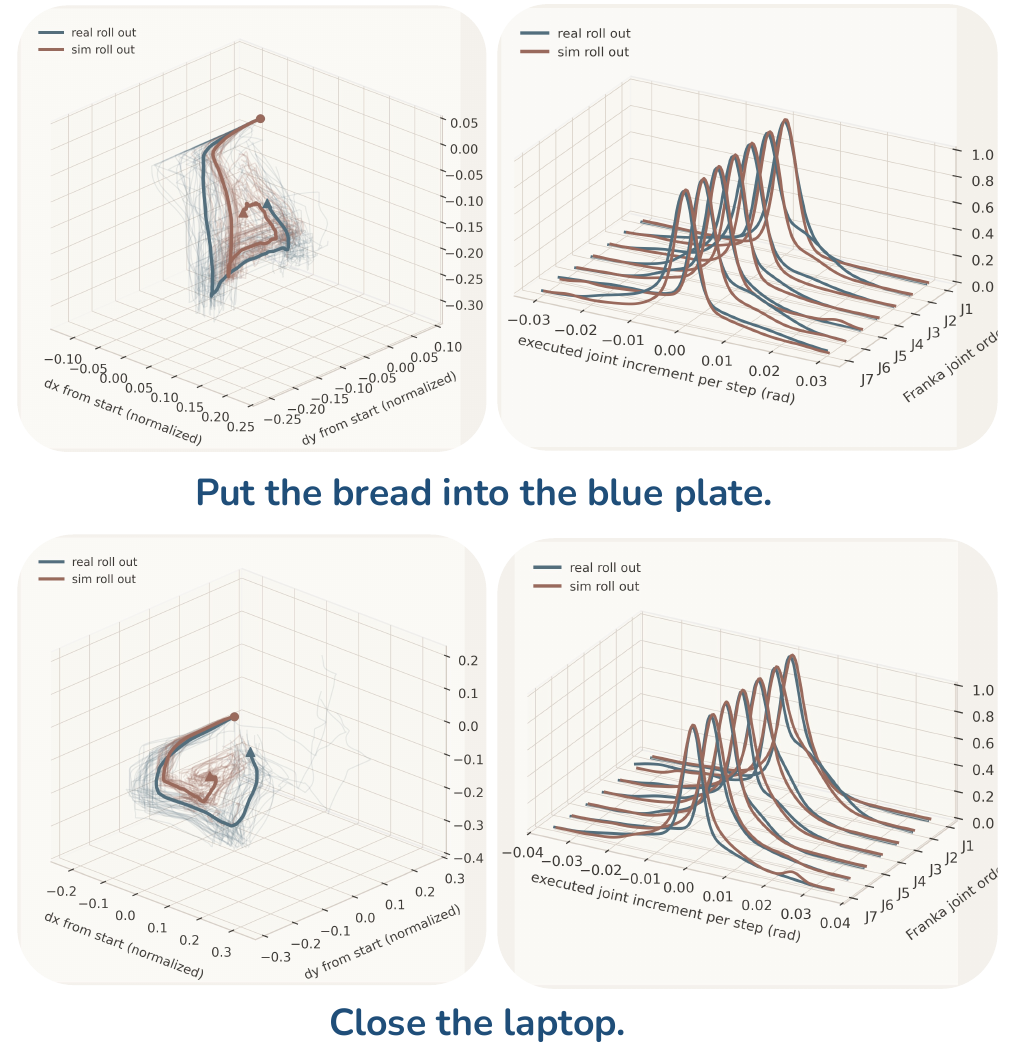}
\vspace{-0.8em}
\caption{\textbf{Behavior-space sim-real comparison.} Left: end-effector displacement trajectories. Right: normalized per-joint distributions of executed joint increments $\Delta q_t^j$.}
\label{fig:gen-eval-behavior}

\end{wrapfigure}

Beyond success-rate correlation, we analyze whether the generated scenes induce similar policy behavior in task space and action space. For each task, we execute the same real-only fine-tuned $\pi_{0.5}$ policy in the real environment and in the corresponding \method-generated simulation scene, without additional adaptation. We record the end-effector trajectory and the executed Franka joint states during each rollout.

For task-space behavior, we express the end-effector trajectory as displacement from its initial position,
\[
\Delta \mathbf{p}_t = \mathbf{p}_t - \mathbf{p}_0,
\]
and normalize the displacement for visualization. This removes global offsets and highlights whether the simulated rollout follows the same interaction-relevant motion pattern as the real rollout. For action-space behavior, we compute the per-step executed joint increments
\[
\Delta q_t^j = q_{t+1}^j - q_t^j,
\]
for each Franka joint \(j\), and compare the normalized marginal distributions of \(\Delta q_t^j\) between real and simulated rollouts.

This diagnostic complements the aggregate sim-real correlation reported in Sec.~4.5. A generated scene may match visual appearance but still induce different behavior if the object layout, robot base, or contact geometry is inaccurate. Similar end-effector trajectories and joint-increment distributions indicate that \method scenes elicit comparable low-level control behavior from real-only trained policy, supporting their use as a generative evaluation proxy for real deployment of manipulation policies.

\section{Method}
\label{app:loss}

\subsection{Layered Scene Reconstruction from a Single Image}
\subsubsection{Background Inpainting}
\label{app:background}

We use Gemini-2.5-flash-image~\citep{nanobanana2025} for background inpainting. Given a masked scene image where the interaction region is removed and only the background is retained, we use the following prompt to recover an empty background while preserving the scene geometry, camera perspective, and lighting.

\begin{promptbox}
    You are an excellent image inpainter. Your task is to inpaint the masked image, where only the background is reserved and the interactive area has been removed.

Please remove all black regions from the scene, since they are not part of the background. Keep the room structure unchanged. Preserve walls, floor, ceiling, windows, and lighting to recover a full background image.
Do not change the camera perspective. Return an empty background with consistent geometry and textures.

I stress again that I am asking you to inpaint the background, not refill the mask area. Make additional careful checks to avoid deleting any remaining background objects. If there are pixels from the interactive area that were not completely removed, remove them as well, leaving only the background.

Special reminder: remove the desktop area as well, since I generate the interaction area separately. I want the background to be empty so that it can hold the generated desk.

\end{promptbox}

\subsection{Simulation-ready Refinement}

\begin{figure}[htbp]
\centering
\includegraphics[width=\linewidth]{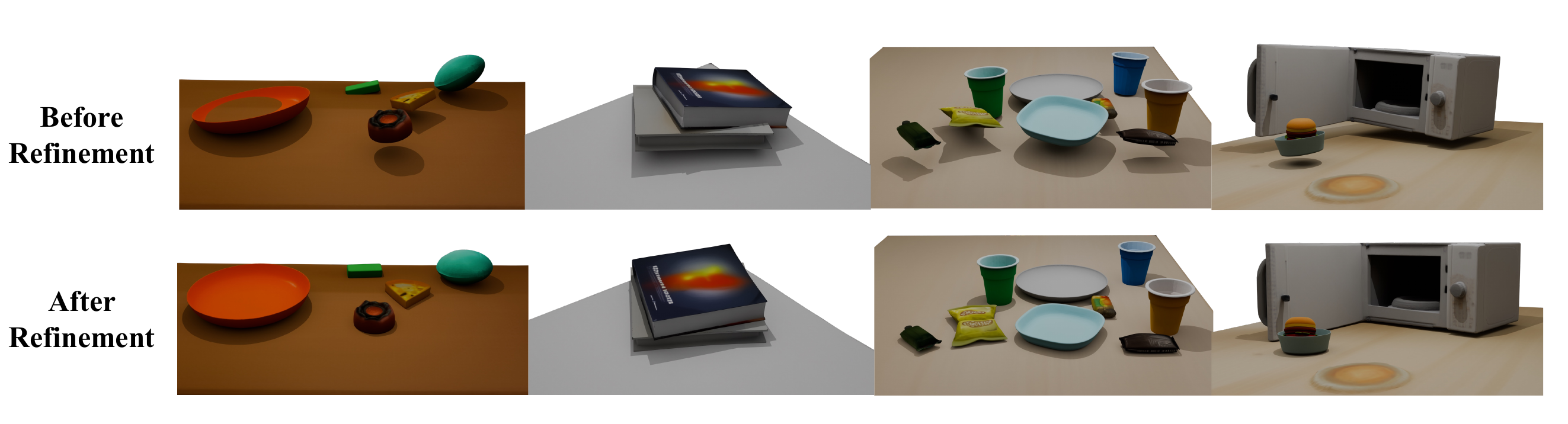}
\vspace{-0.5em}
\caption{\textbf{Visualization of simulation-ready refinement.}
Before refinement, reconstructed object poses can float, interpenetrate, or be unstable. Scene-graph-guided SDF--physics refinement adjusts non-root objects under gravity to satisfy contact constraints, producing physically stable layouts while preserving scene structure.}
\label{fig:droid-sim}
\end{figure}

\paragraph{\textbf{Simulation-ready Refinement Details.}} Algorithm~\ref{alg:alternating} summarizes the full procedure of the Simulation-ready Refinement. Typical hyperparameters are $N_{\text{round}} = 20$, $N_{\text{sdf}} = 15$, $N_{\text{sim}} = 200$, $N_{\text{damp}} = 100$, with convergence threshold $\varepsilon = 10^{-4}$ on maximum object displacement. In the final round, simulation steps are increased $5\times$ to ensure all objects have settled.

\begin{algorithm}[htbp]
\caption{Alternating SDF--Physics Layout Refinement}
\label{alg:alternating}
\begin{algorithmic}[1]
\REQUIRE Meshes $\{\mathcal{M}_i\}_{i=1}^{N}$, initial poses $\{T_i^{\text{init}}\}$, scene graph $\mathcal{G}$
\ENSURE Physically stable world-frame poses $\{T_i\}_{i=1}^{N}$
\STATE Identify roots $\mathcal{R}$ from $\mathcal{G}$ \COMMENT{Supporters never supported}
\STATE \textbf{//} \textit{One-time precomputation}
\FORALL{$i = 1 \ldots N$}
    \STATE Repair $\mathcal{M}_i$ to watertight; compute SDF $\Phi_i$, surface samples $\mathcal{S}_i$
    \STATE Decompose $\mathcal{M}_i$ into convex collision hulls via V-HACD
\ENDFOR
\STATE Build SAPIEN scene with roots $\mathcal{R}$ as kinematic, others as dynamic
\STATE Initialize SDF optimizer with $\{\Phi_i, \mathcal{S}_i, \mathcal{G}\}$
\STATE $\Delta\mathbf{t}_i \gets \mathbf{0},\; \Delta\mathbf{r}_i \gets \mathbf{0}$ for $i \notin \mathcal{R}$; $\{T_i\} \gets \{T_i^{\text{init}}\}$
\FOR{$r = 1$ \TO $N_{\text{round}}$}
    \STATE \textbf{//} \textit{Phase 1: SDF gradient optimization}
    \FOR{$k = 1$ \TO $N_{\text{sdf}}$}
        \STATE Compute $\{T_i\}$ via Eq.~\eqref{eq:delta-pose}
        \STATE $\mathcal{L} \gets w_{\text{pen}}\mathcal{L}_{\text{pen}} + w_{\text{sup}}\mathcal{L}_{\text{sup}} + w_{\text{con}}\mathcal{L}_{\text{con}} + w_{\text{reg}}\mathcal{L}_{\text{reg}}$
        \STATE Adam step on $\{\Delta\mathbf{t}_i, \Delta\mathbf{r}_i\}$ \COMMENT{Freeze $\Delta\mathbf{r}_i$ if $k \le N_{\text{trans}}$}
    \ENDFOR
    \STATE \textbf{//} \textit{Phase 2: Physics settling}
    \STATE Teleport SAPIEN actors to SDF-optimized $\{T_i\}$; reset velocities
    \FOR{$k = 1$ \TO $N_{\text{sim}}$}
        \IF{$k \le N_{\text{damp}}$} \STATE Clamp XY velocities to near-zero \ENDIF
        \STATE Step physics engine
    \ENDFOR
    \STATE Read settled poses $\{T_i^{\text{new}}\}$ from SAPIEN
    \STATE Update SDF optimizer initial poses $\gets \{T_i^{\text{new}}\}$
    \IF{$\max_i \| \mathbf{t}_i^{\text{new}} - \mathbf{t}_i \| < \varepsilon$} \STATE \textbf{break} \ENDIF
    \STATE $\{T_i\} \gets \{T_i^{\text{new}}\}$
\ENDFOR
\RETURN $\{T_i\}$
\end{algorithmic}
\end{algorithm}

We provide the full formulations of the four loss terms used in the SDF phase of Algorithm~\ref{alg:alternating}. For each object $i$, let $\Phi_i(\mathbf{x})$ denote its precomputed signed distance field (negative inside, positive outside) and $\mathcal{S}_i$ its set of $M$ surface sample points, both in the local coordinate frame.

\paragraph{Penetration loss.} For every unordered pair $\{i, j\}$, $i \neq j$, we transform surface samples of $j$ into $i$'s local frame and penalize negative SDF values:
\begin{equation}
\ell_{ij}^{\text{pen}} \;=\; \frac{1}{M} \sum_{\mathbf{s} \in \mathcal{S}_j} \max\!\Bigl(0,\; -\Phi_i\bigl(T_{i \to W}^{-1}\, T_{j \to W}\; \mathbf{s}\bigr)\Bigr), \qquad
\mathcal{L}_{\text{pen}} = \sum_{i \neq j} \bigl(\ell_{ij}^{\text{pen}} + \ell_{ji}^{\text{pen}}\bigr).
\label{eq:pen-loss-detail}
\end{equation}

\paragraph{Support loss.} For each Support edge $(s \to t)$, the supported object $s$ is attracted to the zero-level set of its supporter $t$, with $t$'s pose detached so gradients flow only through $s$:
\begin{equation}
\ell_{st}^{\text{sup}} \;=\; \bigl|\min_{\mathbf{p} \in \mathcal{S}_s} \Phi_t\bigl(T_{t \to W}^{-1}\, T_{s \to W}\, \mathbf{p}\bigr)\bigr| \;+\; \frac{1}{M} \sum_{\mathbf{p} \in \mathcal{S}_s} \max\!\bigl(0,\, -\Phi_t(T_{t \to W}^{-1}\, T_{s \to W}\, \mathbf{p})\bigr).
\label{eq:sup-loss-detail}
\end{equation}
The first term pulls the closest surface point toward contact; the second penalizes penetration. The total is $\mathcal{L}_{\text{sup}} = \sum_{(s \to t) \in \mathcal{E}_{\text{sup}}} \ell_{st}^{\text{sup}}$.

\paragraph{Contact loss.} For each Contact edge $(i, j)$, we apply a symmetric bidirectional penalty:
\begin{equation}
\ell_{ij}^{\text{con}} \;=\; \max\!\bigl(0,\, -\min_{\mathbf{p} \in \mathcal{S}_j} \Phi_i(T_{i \to W}^{-1} T_{j \to W}\, \mathbf{p})\bigr) \;+\; \frac{1}{M} \sum_{\mathbf{p} \in \mathcal{S}_j} \max\!\bigl(0,\, -\Phi_i(T_{i \to W}^{-1} T_{j \to W}\, \mathbf{p})\bigr),
\label{eq:con-loss-detail}
\end{equation}
with $\mathcal{L}_{\text{con}} = \sum_{(i,j) \in \mathcal{E}_{\text{con}}} (\ell_{ij}^{\text{con}} + \ell_{ji}^{\text{con}})$. The first term in each direction prevents separation, the second prevents penetration.

\paragraph{Regularization.} An $\ell_2$ penalty on the residuals prevents drift from the initial estimates:
\begin{equation}
\mathcal{L}_{\text{reg}} \;=\; \sum_{i \notin \mathcal{R}} \Bigl( \|\Delta \mathbf{t}_i\|_2^2 \;+\; \lambda_r \|\Delta \mathbf{r}_i\|_2^2 \Bigr), \quad \lambda_r = 5.
\label{eq:reg-loss-detail}
\end{equation}

\paragraph{SDF precomputation details.}

SDF grids are computed at a resolution of $128^3$ using Open3D’s ray-casting signed distance implementation~\citep{zhou2018open3d} within the axis-aligned bounding box of $\mathcal{M}_i$, expanded by 10\% along each dimension. For non-watertight meshes, we first apply a voxel-based morphological repair procedure: the mesh is voxelized, followed by binary dilation, flood filling from the padded exterior boundary, and a final erosion step to approximately recover the original surface geometry.
Surface samples are constructed from $M = 1024$ uniformly sampled surface points together with all mesh vertices, which helps preserve potential contact regions around sharp geometric features.

\paragraph{Physics simulation details.}

The SAPIEN simulation uses a timestep of $1/100$\,s, a uniform density of $3000$\,kg/m$^3$, friction coefficients of $0.5$ (object--object) and $5.0$ (object--ground), and zero restitution. In each round, all actors are teleported to their SDF-optimized poses, velocities are reset, and the engine is simulated for $N_{\text{sim}}$ steps. During the first $N_{\text{damp}}$ steps, XY velocity magnitudes are clamped to $10^{-7}$ while the $z$ velocity is limited to $0.01$\,m/s, enabling gradual gravity-driven settling without lateral drift and preventing collision impulses from launching objects.

\section{\dataset{} Dataset}
\label{app:droid-sim}
\paragraph{DROID-scale scene construction.}
\method is not restricted to the small subset used for detailed quantitative evaluation. We apply the scene-construction pipeline to DROID at the scene-ID level. DROID~\cite{droid2024} is a large-scale in-the-wild manipulation dataset with 76k demonstrations, 350 hours of interaction, and 564 scenes across diverse tasks and collection sites~\citep{droid2024}. Starting from these DROID scene groups, we construct \dataset{} as a real-to-sim companion dataset. Each recovered scene is linked back to its original DROID raw-data identifier, including the collection-site identifier, success/failure split, collection date, trajectory folder, metadata file, and trajectory files, making the generated simulation assets traceable to the corresponding real demonstration.

\begin{figure}[htbp]
\centering
\includegraphics[width=\linewidth]{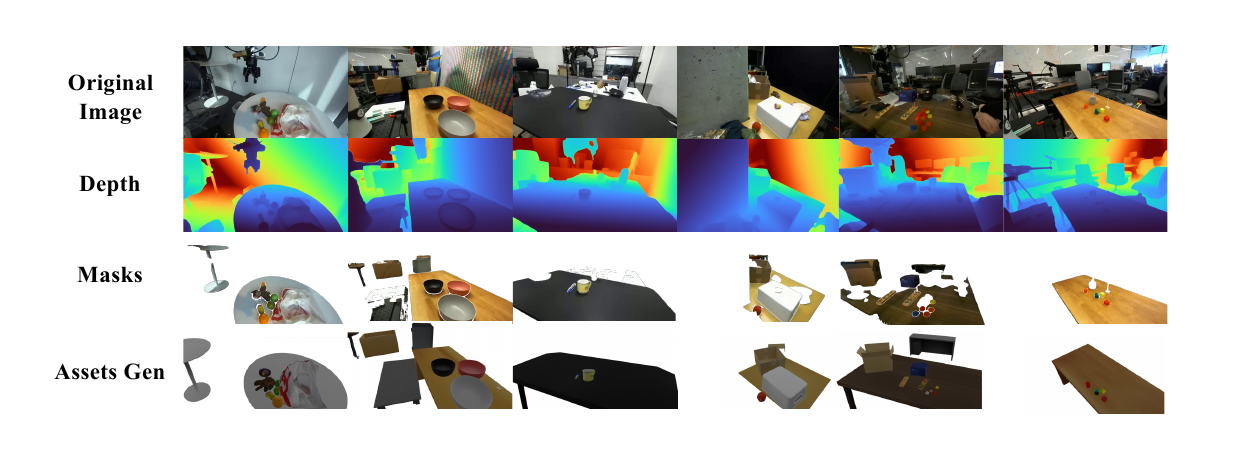}
\vspace{-0.5em}
\caption{\textbf{DROID-scale scene construction.} We parse DROID raw-data identifiers, extract representative RGB frames, generate per-scene object and support-surface prompts, and run open-vocabulary grounding and segmentation before downstream asset generation and simulation refinement.}
\label{fig:droid-sim}
\end{figure}

Our processing is automated after metadata parsing. For each retained scene folder, we extract one RGB frame (usually the first frame of one of the exterior cameras) and query a VLM to produce object and support-surface captions, which are saved as \texttt{mask\_prompt.txt}. We then use Grounding DINO for text-conditioned localization, Florence-2 for referring-expression disambiguation when multiple detections match the same category, and SAM~3 for mask extraction~\citep{liu2023grounding,xiao2023florence,sam3_2025}. 

This design avoids relying on a closed prompt list such as \texttt{can/cup/mug/bowl/box/...} because DROID includes long-tail household objects, tools, appliances, containers, and scene-specific distractors. The generated prompts, annotated views, and merged masks are stored with each recovered scene as an auditable record of the automatic parsing stage.

After foreground mask extraction, we lift each retained object mask into a coarse 3D asset using SAM3D~\cite{sam3d2025} and estimate an initial object pose from the selected RGB frame. We further predict a monocular depth map and reconstruct a masked object point cloud, which is aligned to the SAM3D asset using the initial pose. We then refine the object pose by performing point-to-plane ICP between mesh-sampled surface points and the observed depth point cloud, optimizing only rotation and translation while keeping the scale fixed. For the background, we segment the support surface and static scene region, complete the occluded background, and reconstruct a Gaussian-splatting representation of the scene background.

Based on runtime estimates over several hundred DROID-Sim scenes, the full automated pipeline from a single RGB frame to a foreground-background composited, simulation-ready scene takes approximately 12 to 25 minutes per scene, depending on the number of objects. On average, scenes with moderate complexity (around 5–10 objects) require approximately 17 minutes. The dominant costs come from background completion, Gaussian-splatting reconstruction, and the final simulation-readiness refinement.

\begin{figure}[htbp]
  \vspace{-0.5em}
  \centering
  \includegraphics[width=\linewidth]{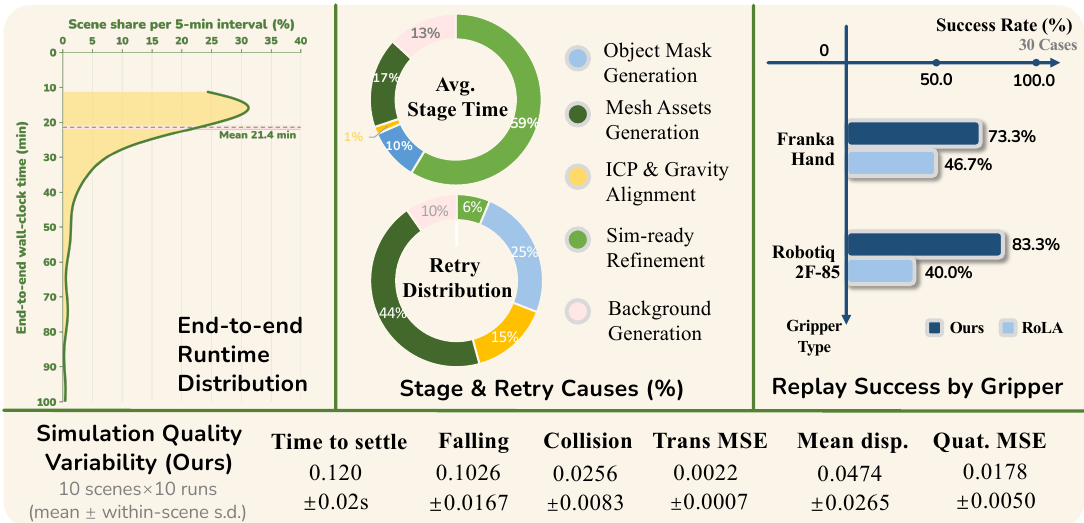}
 \vspace{-0.9em}
  \caption{\textbf{Runtime statistics and gripper-controlled replay results}.%
  \label{fig:statistics}}
  \vspace{-0.5em}
\end{figure}

\paragraph{Runtime and Gripper-Controlled Replay.} We analyze generation logs for all 564 \dataset{} scenes. The average end-to-end runtime is 21.4 minutes per scene, with simulation-ready refinement accounting for 59\% of the pipeline time, primarily due to convex decomposition (Figure~\ref{fig:statistics}). We additionally compare \method{} and RoLA~\citep{rola2025} over 30 replay cases under matched settings using the Franka Hand and Robotiq 2F-85 grippers. \method{} achieves success rates of 73.3\% and 83.3\%, respectively, compared with 46.7\% and 40.0\% for RoLA, showing that its replay advantage persists with both gripper types.

\section{Interactive Scene Construction GUI Tool}
\label{app:gui}

\begin{figure}[htbp]
\centering
\includegraphics[width=\linewidth]{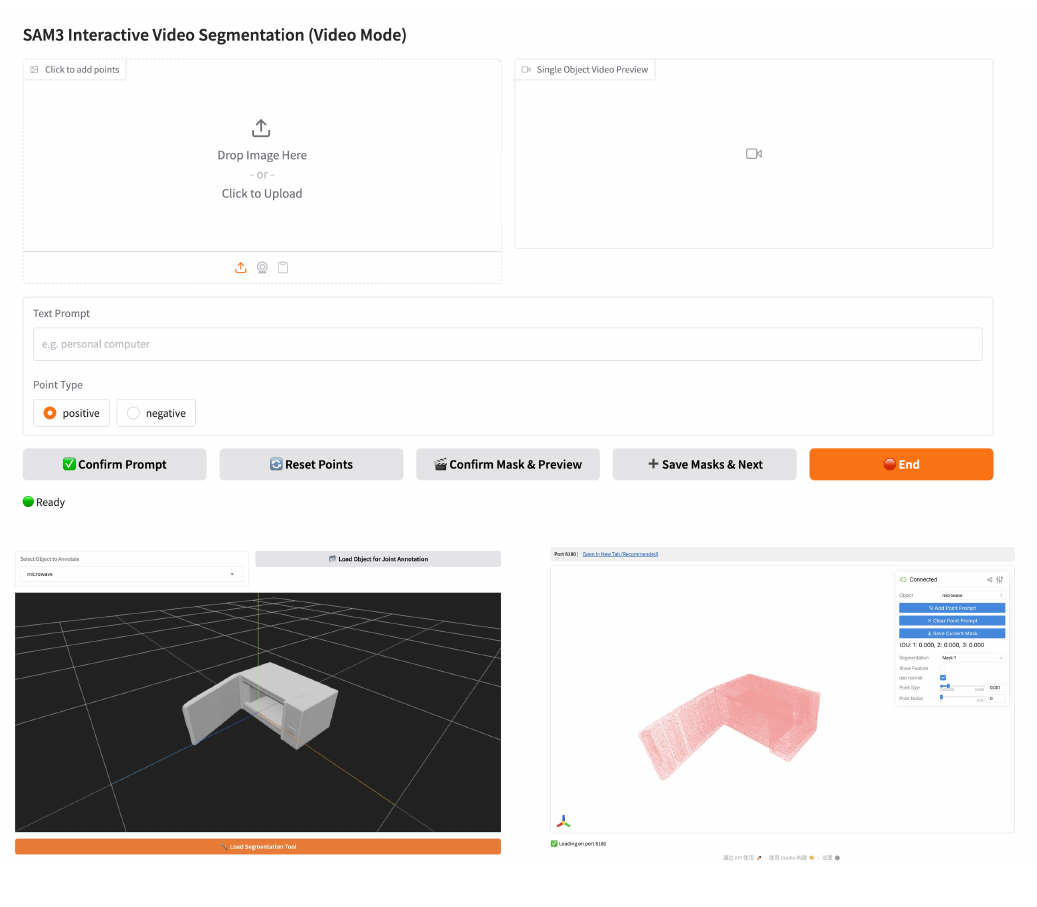}
\vspace{-0.5em}
\caption{
\textbf{Interactive scene construction GUI.}
Our Gradio-based interface covers the full scene construction pipeline from 2D mask annotation to 3D articulated asset preparation.
}
\label{fig:gui-intro}
\end{figure}

We implement an interactive GUI to simplify the construction of articulated 3D scenes from monocular images or videos.
The interface is built with Gradio and integrates the complete workflow, including prompt-based mask initialization, click-based mask refinement, video mask propagation, 3D asset generation, scene-level GLB composition, and articulated-object annotation.
Users first upload an image or video and specify the target object using a text prompt.
The mask can then be refined through positive and negative clicks, allowing users to correct under-segmentation or over-segmentation before saving the object mask.

For video inputs, the system propagates the confirmed mask through the sequence and automatically selects a set of high-quality frames for 3D reconstruction.
This multi-view mask selection improves the quality and completeness of the generated assets compared with relying on a single view.
After all objects are segmented, the GUI invokes the SAM3D-based ~\citep{sam3d2025, mv_sam3d2026} reconstruction pipeline to generate object-level GLB meshes and compose them into a scene-level asset.
The user can then inspect the generated GLB files, select the object of interest, and enter the articulated-object interface, where the selected asset can be further segmented into parts and annotated with articulation information.
This design avoids switching between separate scripts and visualization tools, making the overall scene construction process more reproducible and easier to operate.

\clearpage
\end{document}